\documentclass[12pt,english]{article}

\usepackage{algorithm}
\usepackage{booktabs}
\usepackage{graphicx}
\usepackage{amsmath}
\usepackage{amssymb}
\usepackage{latexsym}
\usepackage{mathtools}
\usepackage{csquotes}
\usepackage{crop}
\usepackage{algorithmic,algorithm}
\usepackage{multirow}
\usepackage{bm}
\usepackage{bbm}
\usepackage{enumerate}
\usepackage{url}
\usepackage{array}
\usepackage{paralist}
\usepackage{diagbox}
\usepackage{wasysym}
\usepackage{booktabs}
\usepackage[dvipsnames]{xcolor}
\usepackage[colorlinks = true, pdfstartview = FitV, linkcolor = blue, citecolor = blue, urlcolor = blue]{hyperref}

\usepackage{listings}

\usepackage{rotating}

\usepackage[capitalise]{cleveref}
\crefname{equation}{}{}
\crefname{figure}{Figure}{Figures}
\creflabelformat{equation}{\textup{(#2#1#3)}}
\crefname{assumption}{Assumption}{Assumptions}
\crefname{condition}{Condition}{Conditions}

\usepackage{xspace}

\usepackage{fullpage}
\usepackage{multirow}

\usepackage[sort,nocompress]{cite}

\usepackage{arydshln}
\usepackage{enumitem}
\setlist[enumerate,1]{leftmargin=*,wide=0em, noitemsep,nolistsep, label = {\bfseries \arabic*.}}
\setlist[itemize,1]{leftmargin=*,wide=0em, noitemsep,nolistsep}

\usepackage{pifont}
\makeatletter
\newcommand*{\transpose}{%
	{\mathpalette\@transpose{}}%
}
\newcommand*{\@transpose}[2]{%
	\raisebox{\depth}{$\m@th#1\intercal$}%
}
\makeatother

\newcommand*\xbar[1]{%
	\hbox{%
		\vbox{%
			\hrule height 0.5pt 
			\kern0.5ex
			\hbox{%
				\kern-0.1em
				\ensuremath{#1}%
				\kern-0.1em
			}%
		}%
	}%
} 

\definecolor{forestgreen}{rgb}{0.13, 0.55, 0.13}

\definecolor{amber}{rgb}{1.0, 0.75, 0.0}

\definecolor{bananayellow}{rgb}{.8, 0.6, 0}

\newcommand{\luke}[1]{{\leavevmode\color{red}{Luke:\ #1}}}

\newcounter{comment}

\usepackage{amsthm}
\usepackage[framemethod=TikZ]{mdframed}

\mdfdefinestyle{theoremstyle}{%
	linewidth = 1pt,%
	roundcorner = 10pt,%
	leftmargin = 0,%
	rightmargin = 0,%
	backgroundcolor = cyan!3,%
	outerlinecolor = magenta!70!black,%
	splittopskip = \topskip,%
	ntheorem = true,%
}
\mdtheorem[style=theoremstyle]{claim}{Claim}

\newmdtheoremenv[%
linewidth = 1pt,%
roundcorner = 10pt,%
leftmargin = 0,%
rightmargin = 0,%
backgroundcolor = green!3,%
outerlinecolor = blue!70!black,%
splittopskip = \topskip,%
ntheorem = true,%
]{theorem}{Theorem}

\newmdtheoremenv[%
linewidth = 1pt,%
roundcorner = 10pt,%
leftmargin = 0,%
rightmargin = 0,%
backgroundcolor = green!3,%
outerlinecolor = blue!70!black,%
splittopskip = \topskip,%
ntheorem = true,%
]{corollary}{Corollary}

\newmdtheoremenv[%
linewidth = 1pt,%
roundcorner = 10pt,%
leftmargin = 0,%
rightmargin = 0,%
backgroundcolor = green!3,%
outerlinecolor = blue!70!black,%
splittopskip = \topskip,%
ntheorem = true,%
]{lemma}{Lemma}

\newmdtheoremenv[%
linewidth = 1pt,%
roundcorner = 10pt,%
leftmargin = 0,%
rightmargin = 0,%
backgroundcolor = yellow!3,%
outerlinecolor = blue!70!black,%
splittopskip = \topskip,%
ntheorem = true,%
]{definition}{Definition}

\newmdtheoremenv[%
linewidth = 1pt,%
roundcorner = 10pt,%
leftmargin = 0,%
rightmargin = 0,%
backgroundcolor = green!3,%
outerlinecolor = blue!70!black,%
splittopskip = \topskip,%
ntheorem = true,%
]{proposition}{Proposition}

\newmdtheoremenv[%
linewidth = 1pt,%
roundcorner = 10pt,%
leftmargin = 0,%
rightmargin = 0,%
backgroundcolor = green!3,%
outerlinecolor = blue!70!black,%
splittopskip = \topskip,%
ntheorem = true,%
]{condition}{Condition}

\newmdtheoremenv[%
linewidth = 1pt,%
roundcorner = 10pt,%
leftmargin = 0,%
rightmargin = 0,%
backgroundcolor = green!3,%
outerlinecolor = blue!70!black,%
splittopskip = \topskip,%
ntheorem = true,%
]{assumption}{Assumption}

\theoremstyle{definition}
\newmdtheoremenv[%
linewidth = 1pt,%
roundcorner = 10pt,%
leftmargin = 0,%
rightmargin = 0,%
backgroundcolor = blue!3,%
outerlinecolor = blue!70!black,%
splittopskip = \topskip,%
ntheorem = true,%
]{example}{Example}

\theoremstyle{definition}
\newmdtheoremenv[%
linewidth = 1pt,%
roundcorner = 10pt,%
leftmargin = 0,%
rightmargin = 0,%
backgroundcolor = red!3,%
outerlinecolor = blue!70!black,%
splittopskip = \topskip,%
ntheorem = true,%
]{remark}{Remark}

\usepackage{tikz}
\usepackage{xparse}

\NewDocumentCommand\DownArrow{O{2.0ex} O{black}}{%
	\mathrel{\tikz[baseline] \draw [<-, line width=0.5pt, #2] (0,0) -- ++(0,#1);}
}

\usepackage{listings} 

\definecolor{mygreen}{rgb}{0,0.6,0}
\definecolor{mygray}{rgb}{0.5,0.5,0.5}
\definecolor{mymauve}{rgb}{0.58,0,0.82}

\usepackage{dsfont}

\allowdisplaybreaks

\newtheorem{observation}{\textbf{Observation}}

\newtheorem{Recommendation}{\textbf{Recommendation}}

\begin{document}
    \title{\Large A Hybrid Statistical-Machine Learning Approach for Analysing Online Customer Behavior: An Empirical Study}
	\author{Saed Alizamir\thanks{School of Management, Yale University, New Haven, CT, USA. Email:  \tt{saed.alizamir@yale.edu}}
		\and
Kasun Bandara\thanks{School of Computing and Information Systems, Melbourne Centre for Data Science, University of Melbourne, VIC, Australia. Email:  \tt{Kasun.Bandara@unimelb.edu.au}}
		\and
Ali Eshragh\thanks{Carey Business School, Johns Hopkins University, MD, and International Computer Science Institute, Berkeley, CA, USA. Email:  \tt{ali.eshragh@jhu.edu}}
		\and
Foaad Iravani\thanks{Uber Freight, San Francisco, CA, USA. Email:  \tt{firavani@uber.com}} 
	}
	\date{}
	\maketitle
	
	
    \begin{abstract}
    	
We apply classical statistical methods in conjunction with the state-of-the-art machine learning techniques to develop a hybrid interpretable model to analyse 454,897 online customers' behavior for a particular product category at the largest online retailer in China, that is JD. While most mere machine learning methods are plagued by the lack of interpretability in practice, our novel hybrid approach will address this practical issue by generating explainable output. This analysis involves identifying what features and characteristics have the most significant impact on customers' purchase behavior, thereby enabling us to predict future sales with a high level of accuracy, and identify the most impactful variables. Our results reveal that customers' product choice is insensitive to the promised delivery time, but this factor significantly impacts customers' order quantity. We also show that the effectiveness of various discounting methods depends on the specific product and the discount size. We identify product classes for which certain discounting approaches are more effective and provide recommendations on better use of different discounting tools. Customers' choice behavior across different product classes is mostly driven by price, and to a lesser extent, by customer demographics. The former finding asks for exercising care in deciding when and how much discount should be offered, whereas the latter identifies opportunities for personalized ads and targeted marketing. Further, to curb customers' batch ordering behavior and avoid the undesirable Bullwhip effect, JD should improve its logistics to ensure faster delivery of orders.

    \end{abstract}



    \newpage
    
    \section{Introduction}\label{sec:introduction}

Accurate sales forecasts are crucial to managing supply chains effectively. Reliable sales forecasts give businesses insights about how they should utilize  resources, control production, inventory, and logistics,  manage risks, and plan for the foreseeable future. Furthermore, accurate sales forecasts drive better performance, and help managers set realistic goals for the business.  

There are three main approaches for forecasting, including classical statistical models \cite{Hyndman2008}, traditional judgmental methods \cite{Lawrence2006}, and state-of-the-art machine learning techniques \cite{Hewamalage2020,Eshragh2022}. While the first two approaches have been extensively developed and studies in the literature \cite{Abolghasemi2020}, the third approach has recently gained popularity \cite{Benidis2020}. 

Statistical predictive models (e.g., linear and logistic regression models) are easy to interpret but typically make some assumptions which might be violated in practice. Alternatively, one may use advanced  machine learning (ML) methods (e.g., neural networks and support vector machines). These sophisticated methods  are more flexible and adaptable to complex (but intuitive) behaviors in reality, such as nonlinear relationships or interactions between explanatory variables. While ML methods can usually provide more accurate predictions, the ``black-box'' nature of these methods makes it difficult to interpret and explain how they generate a prediction or make a recommendation in a way that is understandable by humans. This issue creates a trade-off between accuracy and interpretability \cite{Hastie2009}. 

In a recent review of data analytics applications in operations management, \cite{mivsic2020data} highlight the importance of using interpretable ML models in OM. However, there are only a few recent papers on interpretable models in OM. \cite{ciocan2020interpretable} propose an algorithm for optimal stopping problems, in the form of a binary tree. They show that their tree policies are simple and intuitive and outperform the non-interpretable ML methods. \cite{bravo2020mining} develop an ML approach, called \textit{mining optimal policies} (MinOP), that extracts the structural properties of optimal policies for stochastic dynamic programs from numerical solutions to problem instances. \cite{aouad2019market} provide a decision-tree framework for segmenting users in a population-based on differences in their response patterns.  

Explainable AI (stands for Artificial Intelligence) (or Trusted AI) is an emerging development within AI that proposes methods to help interpret the predictions of a complex model, while maintaining the high level of prediction accuracy \cite{molnar2020interpretable}. The main objective of Explainable AI is to build the trust of users in a model by  offering  transparent reasons for why the model makes a prediction. The ML methods are used for decision making in a wide range of industries, such as manufacturing, healthcare, education, public policy, and finance. Users will be reluctant to accept and act upon model outcomes if they cannot understand how the model makes recommendations, especially if the consequences can be catastrophic (e.g., medical diagnosis or counter-terrorism). Moreover, regulators are increasingly passing legislation (e.g., the General Data Protection Rules in the European Union), requiring companies to provide an explanation of their data-driven decisions to consumers \cite{doshi2017towards, goodman2017european}. Currently, the state-of-the-art algorithm in Explainable AI is SHAP (SHapley Additive exPlanations) developed by \cite{Lundberg2017}. The SHAP framework has proved to be an important advancement in the field of machine learning model interpretation and provides an intuitive unified  approach to explaining the output of any ML model. 

Our paper makes three contributions to the burgeoning field of ML models in OM. First, we use a highly-efficient recent algorithm (LightGBM) to predict JD's sales. We show that LightGBM performs better than LASSO Regression, Ridge Regression and KNN Regressor. Second, we use SHAP   to interpret these predictions. Using SHAP's global interpretability property, we show the importance of each variable in predicting sales. We also present individual SHAP value plots that describe each variable's marginal contribution to the predicted purchase volume of any individual user. Third, we present some of the significant interaction terms that are identified by SHAP. 

To the best of our knowledge, this work is the first attempt in the OM community to use modern ML algorithms to predict online users' purchase behavior and, more importantly, offer clear and insightful interpretations about the prediction model's outcomes. Our analysis helps JD managers identify features/attributes that have the highest impact on sales volume and the direction of the association between these features/attributes and sales. Using this information, the managers can increase sales volume by focusing their efforts and investments on the most impactful variables. When our analytical framework is used to predict a given user's total purchase, the managers can clearly measure and interpret the portion of the forecasted sales that is attributable to each variable. In other words, our analysis helps managers explain \textit{why} an individual prediction is obtained. 

The rest of the paper is organized as follows. Section \ref{sec:literature} reviews the literature. Section \ref{sec:data} describes the data. Section \ref{sec:sales} analyzes product sales  and Section \ref{sec:class} analyzes  customer choice behavior.  We conclude the paper with a discussion in Section \ref{sec:conclusion}.

    \section{Literature}
\label{sec:literature}


\textit{Forecasting} is the art of utilizing the available data to predict the future values of a dynamic system evolving in time. It has a wide range of applications spanning from business and health to science and engineering 
\cite{Cha2004,She2018,Eshragh2019,Messner2019,Abolghasemi2020,Eshragh2020}. 
In particular, in supply chain management (SCM), product demand forecasting plays a crucial role as numerous decisions such as production planning, logistics, and inventory management heavily rely on demand forecasts \cite{Fildes2019}. Therefore, providing accurate demand forecasts can directly result in better operational efficiency, customer satisfaction, and financial savings throughout the entire supply chain \cite{Kremer2015,Trapero2015}. 

In general, there are three main approaches for demand forecasting as listed and explained below: 
\begin{description}
	\item[(i) Statistical Models.] Statistical forecasting models are the time series models that use merely historical data to predict   future values. Some common statistical forecasting models include (but are not limited to) exponential smoothing, and auto-regressive integrated moving average (ARIMA) model (also known as Box-Jenkins model) along with its numerous variants (e.g., SARIMA, ARIMAX, and GARCH) \cite{Hamilton1989,Eshragh2021}.
	
	\noindent In the context of SCM, statistical forecasting models have been developed with advancements in computational power, software and information system technologies such as enterprise resource planning systems, electronic data interchange, and point of sale scanning \cite{Sanders2003}. Such advancements have enabled vast amounts of data to be easily collected, analyzed, and shared throughout the supply chain. A more advanced form of these models are hybrid time series-regression models which incorporate the historical data of the primary time series (e.g., demand) along with one or more explanatory independent variables influencing the primary time series (e.g., promotions, holidays and special events that can impact customer demand) \cite{Trapero2013,Abolghasemi2020}. 
	
	\noindent Although, in practice, there exist several potential candidates for explanatory independent variables, it is prudent to keep models as parsimonious as possible while maintaining desired accuracy \cite{Charkhgard2019}. This is because implementing hybrid time series-regression models with a high number of explanatory variables would be costly in terms of time and memory. Therefore, they are not frequently utilized in industry due to high costs, lack of internal expertise and resources, as well as other organizational barriers \cite{Trapero2015}.

	\item[(ii) Judgmental Methods.] Judgmental forecasting methods refer to traditional techniques incorporating intuitive judgment, opinions and subjective probability estimates provided by experts. Although it has been shown that there are usually high biases   in forecasts provided by human \cite{Tversky1974}, there has been recent increase in integrating these methods with statistical models so that complementary benefits can be realized to alleviate the drawbacks of each approach \cite{Alvarado-Valencia2017,Baecke2017}. 
	
	\noindent \cite{Lawrence2000} classify the useful data for demand forecasting into two classes: (1) historical data, and (2) contextual knowledge. While the former data is simply the recorded time series of historical product sales, contextual knowledge is any other information relevant to interpreting, explaining and anticipating time series behavior, such as promotional plans, competitor activities, and sudden climate changes. Statistical methods are well suited to handle vast amounts of historical data, but when the effects of discontinuities which are often caused by contextual factors, cannot be estimated from historical data, human judgment can be utilized to overcome this issue and incorporate valuable contextual information by adjusting baseline statistical forecasts \cite{Goodwin2002,Lawrence2006,Kremer2015}. 
	The demonstrated accuracy improvements through judgmental forecast adjustments have inevitably resulted in its widespread use in industry \cite{Sanders2003,Fildes2009,Moritz2014}.

	\item[(iii) State-of-the-art Machine Learning Techniques.] Machine learning techniques are increasing in popularity with applications in many different areas. In particular, data scientists have applied these state-of-the-art methods to develop novel models and algorithms for time series forecasting. In this new approach, historical data are used to train a (deep) neural network (NN) to predict the future values. Due to the dependency structure of time series data, recurrent neural networks (RNNs) are the most common class of NNs used in forecasting \cite{Hewamalage2020}. A few packages have been developed for utilizing RNNs in predicting time series data \cite{Alexandrov2019}. 
	
	\noindent In supply chain demand forecasting problems, the requirement is often to produce forecasts for several product demands  that may have similar patterns. In such scenarios, global models can learn across input time series to incorporate more information, in comparison to local methods \cite{Trapero2015}. Examples include 
	\cite{Mukherjee2018,Bandara2019,Salinas2020}. For a more comprehensive literature review  refer to \cite{Benidis2020}.
	
	\noindent Although ML methods can provide accurate forecasts, unlike statistical models, they may not be easily explained and interpreted. The address this major drawback of ML methods, Explainable AI has been emerged. A novel algorithm in this area is called SHAP \cite{Lundberg2017}. The SHAP algorithm provides an intuitive   approach to interpreting the output of ML models. The idea of SHAP  comes from the classic  Shapley values that determine the contribution of each player in a cooperative game \cite{shapley1953value}. The SHAP values quantify the contribution that each feature (variable)\footnotemark \footnotetext{We will be using the terms ``variable'' and ``feature'', interchangeably.} brings to the prediction made by a  model. The SHAP values has three important properties:
	\begin{enumerate}
		\item \textit{Global Interpretability}: SHAP values collectively can show how much each predictor variable contributed to the target variable, either positively or negatively.
		
		\item \textit{Local Interpretability}: Each prediction receives its own set of SHAP values. This is a useful property that significantly enhances the interpretability of ML algorithms.  Traditional algorithms that analyze variable importance only measure the results over the entire population, but not on individual predictions. Local interpretability enables users to compare the impacts of different features across different predictions. 
		
		\item \textit{Computability}: SHAP values can be calculated for any tree-based model, whereas other methods use linear or logistic regression models. 
	\end{enumerate}
\end{description}

In this paper we mainly apply statistical methods and ML techniques to analyze JD's data. To provide interpretability to our ML outcome, we utilize the SHAP algorithm in our analysis. The details of our analysis are elaborated in Sections \ref{sec:data}, \ref{sec:sales}, and \ref{sec:class}. 


    \section{JD Data} \label{sec:data}

We analyze JD.com's user transaction data for March $2018$. The MSOM Society provided the data for the $2020$ Data-Driven Challenge. 
JD  is China's largest online retailer and its biggest retailer overall. A member of the Fortune Global $500$, JD was founded in $1998$ and its retail platform moved online in $2004$. JD's logistic network provides standard same- and next-day delivery of a wide range of products to $99\%$ of China's population (\url{www.JD.com}). This section provides data description as well as empirical statistical analysis on users spending behavior.

\subsection{Data Description}
\label{sec:data_description}

The dataset consists of $20.2$ million click records from $2.56$ million known users who explored $31,868$ SKUs from one product category during the month of March in $2018$. Out of these numbers, only $454,897$ users (i.e., $17.78\%$ of known users) ordered at least one item during the month. Furthermore, these $454,897$ buyer users consists of $374,124$ users who explored the site before placing their order (i.e., there exist some click records for each of them before placing an order), $20,573$ who appear to have made their purchase decisions before logging in (no pre-order interaction data is available), and $60,200$ users for whom we have no interaction information. Thus, in total $80,773$ users' first interaction with the site was an order (i.e., $17.76\%$ of the total buyer users).

It should be noted that although there are $457,298$ users in the data file \texttt{JD\underline\ user\underline\ data.csv}, the order data indicates that only $454,897$ users ordered at least one item during March $2018$ and $2401$ users in the users data file did not place any order during this month.

\noindent \textbf{Number of orders.} Table \ref{tab:number_of_orders_by_buyers} displays the number of orders made by buyer users in terms of both the actual number and percentage. More specifically, each column shows the number of buyer users and the corresponding percentage for the given number of orders varying from $1$ to $9$ units. The last column labeled by $10+$ includes the orders of $10$ units or more. It is readily seen that the majority of buyer users, that is $430,743$ users comprising $94.69\%$ of the total buyer users, ordered only $1$ unit. Note that the largest order size made by a user was $521$ units.  
\begin{table}[tbh!]
	\centering
	\begin{tabular}{|l|c|c|c|c|c|c|c|c|c|c|c|}\hline
		\textbf{Number of orders} & $1$ & $2$ & $3$ & $4$ & $5$ & $6$ & $7$ & $8$ & $9$ & $10+$ \\\hline 
		\textbf{Number of buyers}  & $430,743$ & $21,688$ & $1,832$ & $284$ & $99$ & $42$ & $37$ & $22$ & $19$ & $131$ \\\hline 
		\textbf{Percentage $\%$}  & $94.69$ & $4.77$ & $0.40$ & $0.06$ & $0.02$ & $0.01$ & $0.01$ & $< 0.01$ & $< 0.01$ & $0.03$ \\\hline 
	\end{tabular}
	\vspace{0.3 cm}
	\caption{Number of orders made by buyer users}
	\label{tab:number_of_orders_by_buyers}
\end{table} 

\begin{figure}[h!]
	\centering
	\includegraphics[angle=0, scale=0.45]{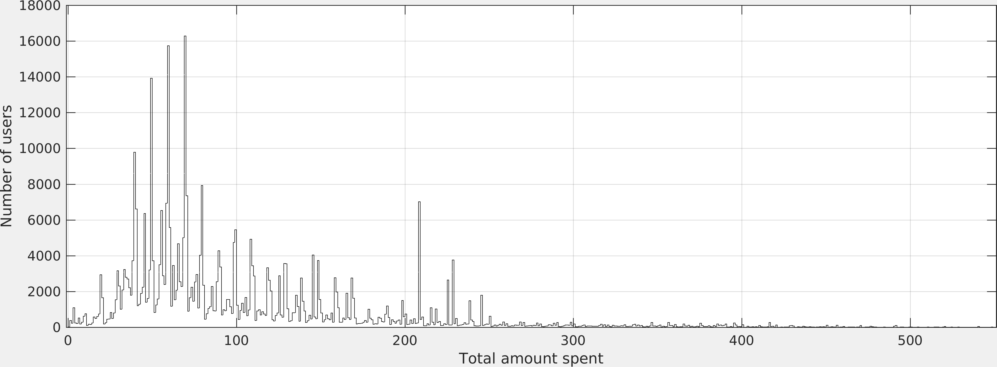}
	\caption{Histogram of amount of money (in RMB) spent by buyer users in March $2018$}
	\label{fig:histogram-spend_per_ordering_user}
\end{figure}

\begin{figure}[h!]
	\centering
	\includegraphics[angle=0, scale=0.45]{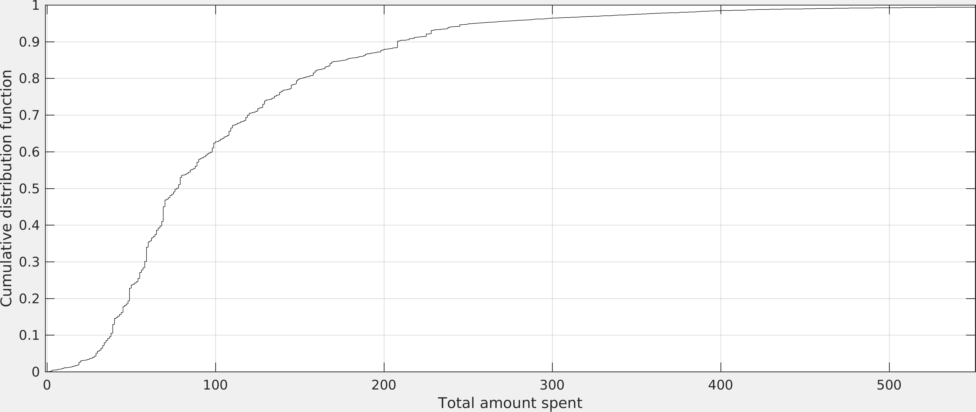}
	\caption{Cumulative distribution function of amount of money (in RMB) spent by buyer users in March $2018$}
	\label{fig:CDF-spend_per_ordering_user}
\end{figure}

\noindent \textbf{Total spent money.} Figures \ref{fig:histogram-spend_per_ordering_user} and \ref{fig:CDF-spend_per_ordering_user} display the histogram and cumulative distribution function of the total amount of money (in RMB) spent by buyer user in March $2018$. Figure \ref{fig:histogram-spend_per_ordering_user} shows that the distribution of the total amount spent by users is skewed to the right with a mode around RMB $60$ (after smoothing out the plot). Furthermore, Figure \ref{fig:CDF-spend_per_ordering_user} illustrates that while the median (i.e., $50\%$ percentile) of the total amount spent by buyer user is RMB $80$, around $90\%$ of buyer users spent at most RMB $210$ in this product category in March $2018$. 

\begin{figure}[h!]
	\centering
	\includegraphics[angle=0, scale=0.45]{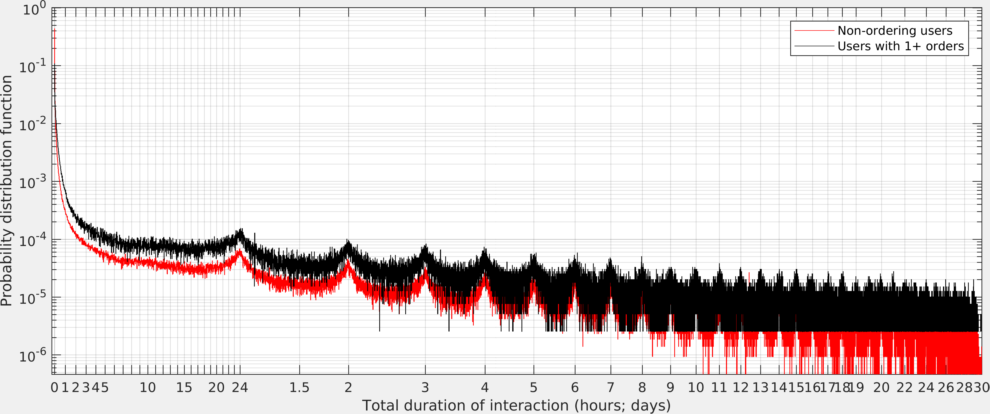}
	\caption{Probability distribution function of the total duration of interaction}
	\label{fig:PDF-duration_of_interaction}
\end{figure}

\begin{figure}[h!]
	\centering
	\includegraphics[angle=0, scale=0.45]{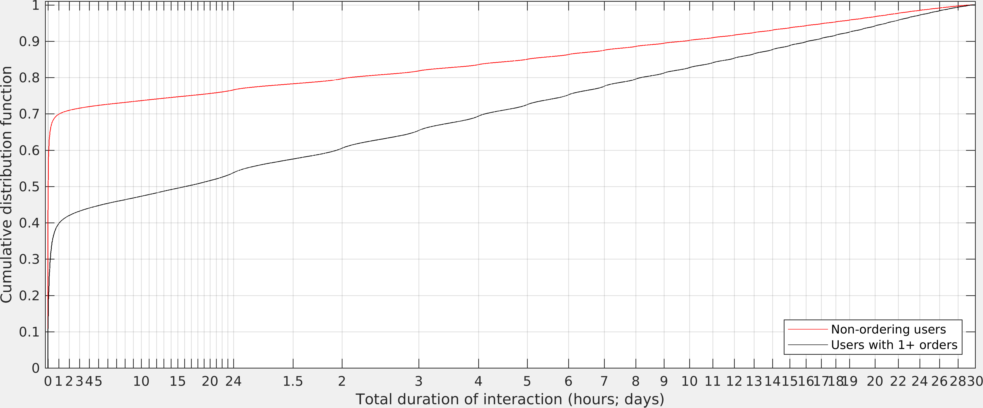}
	\caption{Cumulative distribution function of the total duration of interaction}
	\label{fig:CDF-duration_of_interaction}
\end{figure}

\noindent \textbf{User interaction time.} Figures \ref{fig:PDF-duration_of_interaction} and \ref{fig:CDF-duration_of_interaction} display the probability and cumulative distribution function of \textit{user interaction time} with the site for both buyer and non-buyer users (via whichever channel), respectively. More specifically, they show the time elapsed between a user's first and last record over the whole month. A typical user interaction involves a few clicks on different SKUs which may result in ordering one of them. In both figures, the bin size is $10$ minutes and the x-axis is log scale. The x-axis labels are hours on the left varying from $1$ to $24$ hours, and then become days covering all $30$ days. Such scaling illustrates the detail of early times while still spreading out the rest of the month enough. Figure \ref{fig:PDF-duration_of_interaction} exhibits a $24$-hour frequency component to the buyer users' visits, that is people were often coming back to the site at about the same time each day. Moreover, Figure \ref{fig:CDF-duration_of_interaction} reveals that $40\%$, $54\%$ and $78\%$ of buyer users had less than $1$ hour's, $1$ days', and $1$ week's interaction with the site, respectively. These percentages for non-buyer users are $70\%$, $77\%$ and $88\%$, respectively. This implies that in average, buyer users spend more time on the site comparing with non-buyer users.

Although the click data are bearing valuable information for making predictions, considering them as independent samples may not be an accurate approach. Because there are some obvious significant correlations between multiple clicks of a buyer user led to an order. To address this issue, for each buyer user, we eliminate any click activity that they had before placing their order and keep only the last click record on the SKU in which they ordered. However, $80,773$ buyer users' whose first interaction with the site was an order did not have any pre-click history. For these users, we keep their order records as their successful clicks. Although the click records include the \texttt{click\underline{ }channel} data, the order records do not provide this information. Therefore, for these buyer users, we introduce a new channel, namely \texttt{unknown}.

\subsection{Empirical Statistical Analysis on Users' Spending Behavior}
\label{sec:data_empirical_sa}

\begin{figure}[h!]
	\centering
	\includegraphics[angle=0, scale=0.45]{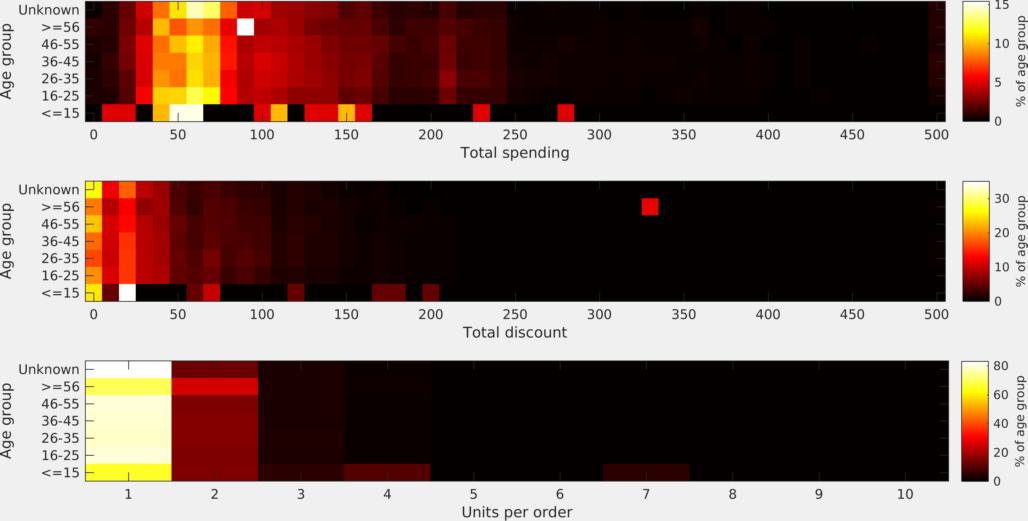}
	\caption{The distribution of total spending money, total discount, and number of orders vs. age groups}
	\label{fig:field-age_effects}
\end{figure}

\begin{figure}[h!]
	\centering
	\includegraphics[angle=0, scale=0.45]{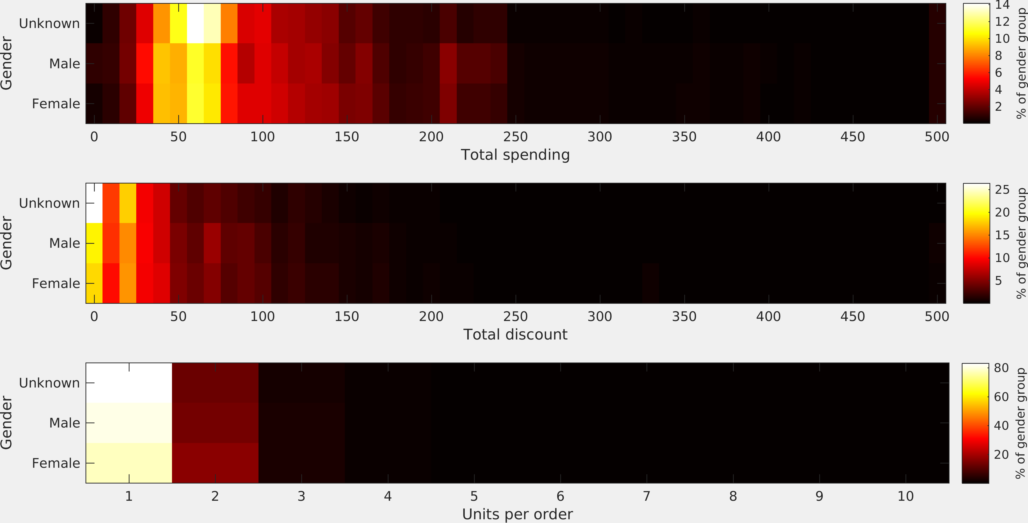}
	\caption{The distribution of total spending money, total discount, and number of orders vs. gender}
	\label{fig:field-gender_effects}
\end{figure}

\begin{figure}[h!]
	\centering
	\includegraphics[angle=0, scale=0.45]{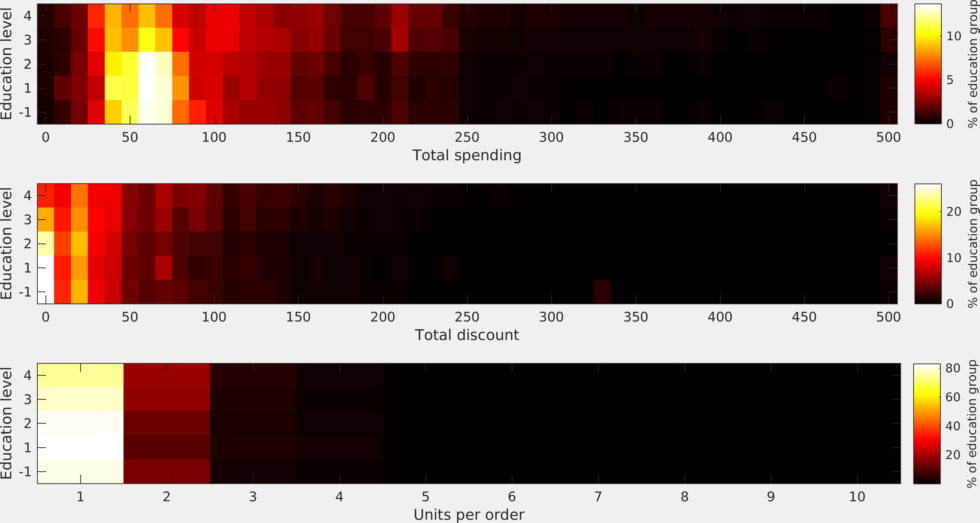}
	\caption{The distribution of total spending money, total discount, and number of orders vs. education levels}
	\label{fig:field-education_effects}
\end{figure}

\begin{figure}[h!]
	\centering
	\includegraphics[angle=0, scale=0.45]{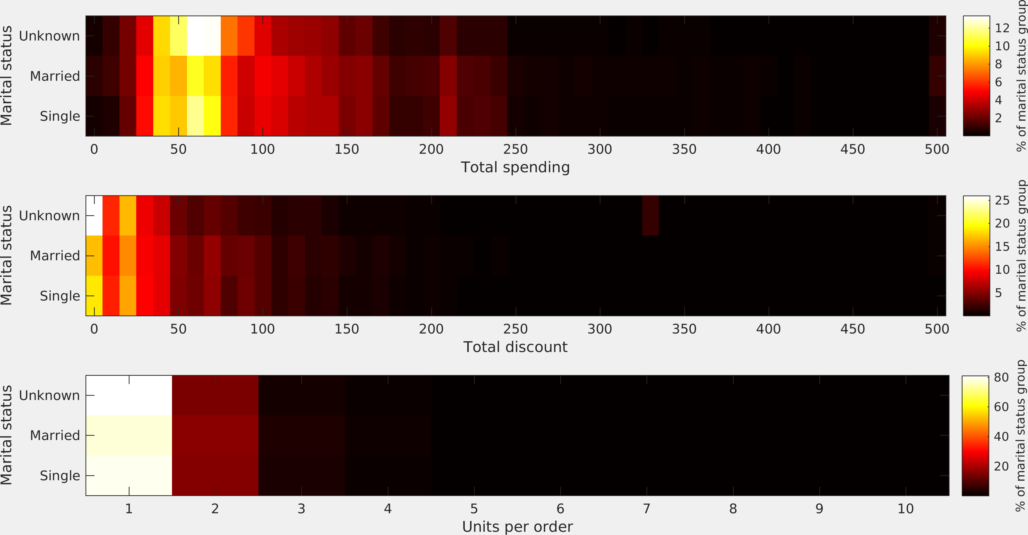}
	\caption{The distribution of total spending money, total discount, and number of orders vs. marital status }
	\label{fig:field-maritalstatus_effects}
\end{figure}

\begin{figure}[h!]
	\centering
	\includegraphics[angle=0, scale=0.45]{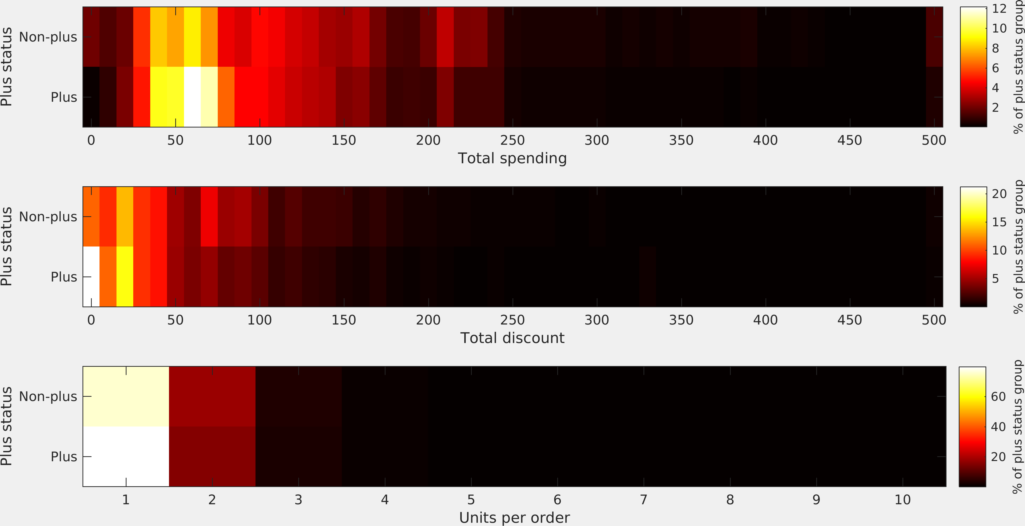}
	\caption{The distribution of total spending money, total discount, and number of orders vs. plus status levels}
	\label{fig:field-plusstatus_effects}
\end{figure}

\begin{figure}[h!]
	\centering
	\includegraphics[angle=0, scale=0.45]{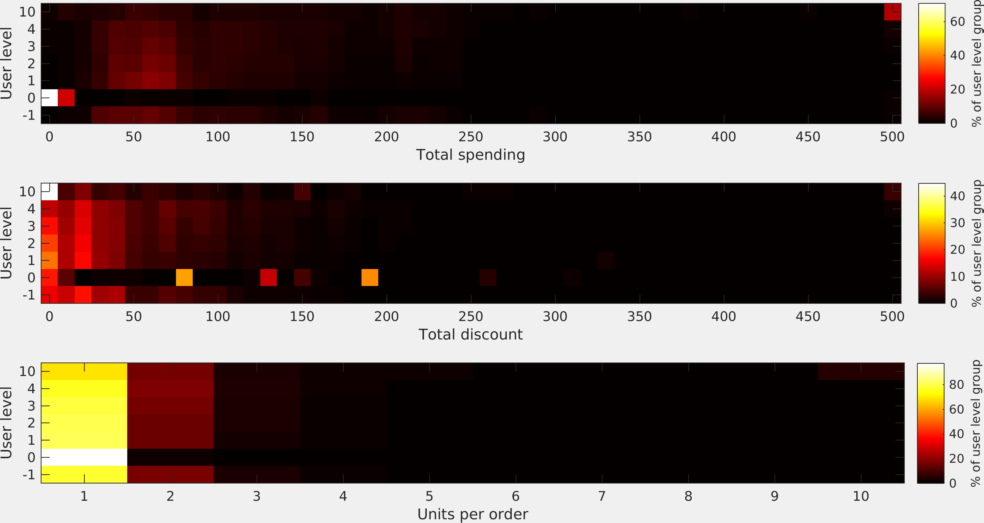}
	\caption{The distribution of total spending money, total discount, and number of orders vs.  user levels}
	\label{fig:field-userlevel_effects}
\end{figure}

This section studies the impact of six categorical features consisting of ``Age'', ``Gender'', ``Education'', ``Marital status'', ``Plus status'', and ``User level'' on three numerical variables including total spending money by buyer users, total discount that buyer users received, and total number of units that buyer users ordered. For this purpose, we construct the empirical probability distribution for each level of the given categorical variables and display them in two-dimensional histograms, with each row is normalized, so the sum across rows is $100\%$. In all Figures \ref{fig:field-age_effects}--\ref{fig:field-userlevel_effects}, the lighter/darker colors imply the higher/lower percentage. In addition, total spending and total discount bins are $0$ to $500$ in steps of $10$, and units per order bins are $1$ to $10$. The last bin labeled $10$, includes all orders of 10 units or more. Furthermore, the level ``Unknown" refers to those buyer users whose values for the corresponding features are not provided in the dataset. 

\noindent \textbf{Age.} The feature age is categorized into seven levels consisting of ``less than $16$'', ``$16-25$'', ``$26-35$'', ``$36-45$'', ``$46-55$'', ``greater than $55$'', and ``Unknown''. Figure \ref{fig:field-age_effects} indicates that for all age groups varying between $16-55$ as well as the unknown group, their total spending distributions are similar, achieving their spike at RMB $60$. However, the youngest age group (i.e., less than $16$) total spending distribution has an exotic pattern with a mode at RMB $60$ and several peaks, and the oldest age group (i.e., greater than $ 55$) total spending distribution achieves its spike at RMB $90$. A similar behavior appears for the total discount distribution. While the age groups between $16-55$ as well as the unknown group attain their maximum at RMB zero, the youngest group reaches its maximum at RMB $20$ and the oldest age group has two nice peaks one at RMB zero and the other one at RMB $340$. The behavior of buyer users in all age groups for the order volume seem to be similar achieving their peaks at one unit.

\noindent \textbf{Gender.} The feature gender is categorized into three levels consisting of ``Female'', ``Male'', and ``Unknown''. Figure \ref{fig:field-gender_effects} shows that the distribution of total spending , total discount, and order volume for all three gender groups are skewed to the right with peaks at RMB $60$, RMB zero, and one unit, respectively. Furthermore, the total spending distribution of the male and female groups achieves a second spike at RMB $210$. It appears that the unknown group spends more than the other two gender groups in average. 

\noindent \textbf{Education.} The feature education is categorized into five levels consisting of ``$1$: less than high school'', ``$2$: high school diploma or equivalent'', ``$3$: Bachelor’s degree'', ``$4$: post-graduate degree'', and ``$-1$: unknown''. Figure \ref{fig:field-education_effects} shows that the total spending distribution for all education groups is skewed to the right and achieves its spike at RMB $60$, but the highest two education groups $3$ and $4$ posses a couple of further peaks at RMB $100$ and RMB $210$. Furthermore, while the education groups $-1$, $1$ and $2$ have a heavier low spending tail, the last two education groups $3$ and $4$ have a longer high spending tail. It appears that the two education groups $1$ and $2$ spend more than the other groups, in average. Similar to the previous features, most discounts are small (peaked at zero for all education groups) and pretty much everyone orders only one unit at a time. 

\noindent \textbf{Marital status.} The feature marital status is categorized into three levels consisting of ``Single'', ``Married'', and ``Unknown''. Figure \ref{fig:field-maritalstatus_effects} shows that the total spending distribution for all marital status groups is skewed to the right and achieves its spike at RMB $60$, but the two single and married groups posses one further peak at RMB $210$ and have a longer high spending tail. In average, the unknown group spends more than the other two marital status groups. Similar to the previous features, most discounts are small (peaked at zero for all marital status groups) and pretty much everyone orders only one unit at a time. 

\noindent \textbf{Plus status.} The feature plus status is categorized into two levels consisting of ``Plus'' and ``Non-plus''. Figure \ref{fig:field-plusstatus_effects} shows that the total spending distribution for both plus status groups is skewed to the right and achieves its spike at RMB $60$. While the plus group has a smaller variance and lighter tails, the non-plus group has a flatter distribution with heavier tails. In addition, the plus group spends more than the non-plus group, in average. Unlike the previous features, while the total discount distribution for the plus group achieves its maximum at RMB zero and a second peak at RMB $20$, the corresponding distribution for the non-plus group achieves its maximum at RMB $20$ with a second peak at RMB $70$. This implies that discount is more enticing for non-plus users than the plus users. Again, in consistent with the previous results, pretty much everyone orders only one unit at a time.

\noindent \textbf{User level.} The feature user level is categorized into seven levels consisting of ``$-1$'', ``$0$'', ``$1$'', ``$2$'', ``$3$'',  ``$4$'', and ``$10$'', where $-1$ corresponds to new buyer users, $0-4$ correspond to existing users where a higher user level is associated with a higher total purchase value in the past, and $10$ corresponds to enterprise users. Figure \ref{fig:field-userlevel_effects} shows that the total spending distribution for all user levels $-1$ and $1-4$ is skewed to the right, has light tails, with the maximum achieved at RMB $60$, the zero user level has a very low variance with the maximum achieved at RMB zero, and the user level $10$ is flat with a big spike at RMB $500$. Furthermore, while the user level $-1$ has a heavier low spending tail, the user levels $1-4$ have a longer high spending tail. In addition, buyer users in level $0$ spends more than the other levels, in average. Such a similar pattern is observed for the total discount. The total discount distribution for all user levels $-1$ and $1-4$ has two peaks at RMB zero and RMB $60$, the zero user level has a few peaks at RMB zero, RMB $80$, RMB $130$, and RMB $190$, and the enterprise users (i.e., user level $10$) has a maximum at $0$ level. This observation implies that while the behavior of new users and more established individual users are similar, those users who have recently purchased through the site for the first time (i.e., user level $0$) behave more sensitively to the discounts offered by the site. As expected, the enterprise users are insensitive to discounts. Regarding the order volume distribution, while most individual users order only one unit at a time, the corresponding distribution for the enterprise users attains two spikes at $1$ unit and $10+$ units.
\vspace{-0.5cm}

\begin{figure}[h!]
	\centering
	\includegraphics[angle=0, scale=0.45]{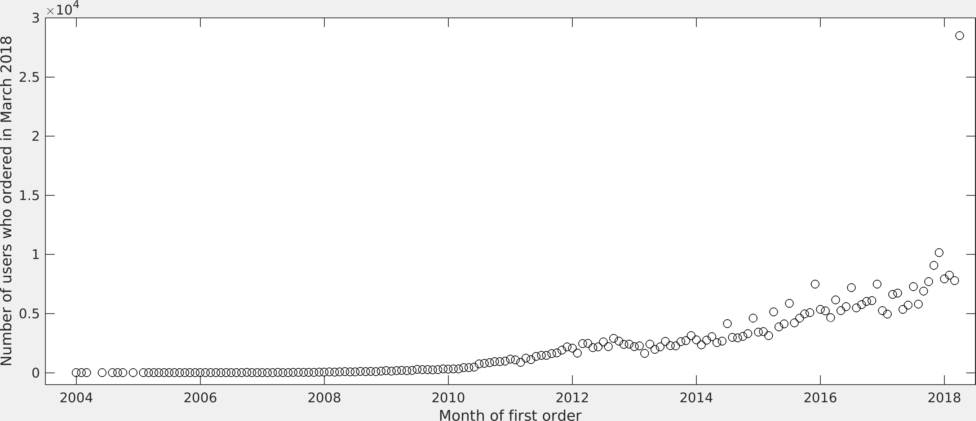}
	\caption{The first order placed by buyer users in March $2018$}
	\label{fig:first_order_month}
\end{figure}

Figure \ref{fig:first_order_month} shows the month of the first order placed by buyer users in March $2018$. It reveals that while the oldest month of the first order placed by a user, was December $2003$, there has been an increasing trend in the number of new users such that $28,497$ users (i.e., $6.26\%$ of the total $454,897$ buyer users in this month) made their first order in March $2018$. These are users with level $-1$. 

\begin{observation}
	In summary, while the spending behavior of buyer users with respect to the features age, gender, education, and marital status appear similar, their behavior with respect to the two features plus status and user levels are totally different. Such observation is made for the total discount distribution as well. However, irrespective of the chosen feature, the majority of buyer users order only one unit at a time which is consistent with the results reported in Table \ref{tab:number_of_orders_by_buyers}.
\end{observation}

    \section{Sales Volume Analysis} \label{sec:sales}
In this and the  next sections, we examine 
JD customers'   behavior   and identify the most important factors that influence their  purchase behavior. In this section,  we build a model to predict customers' purchase volume (sales). In Section \ref{sec:class},
we build a classification model to explore customers' product choice decision.  We now describe the steps   we took to build our sales prediction model.    

\subsection{Variable Selection}

The first  step of our analysis was choosing the explanatory variables for predicting sales.  
After  reviewing the data  , we decided to use the following     explanatory variables.

\subsubsection{Customer Variables:} 
We use Gender, Marital Status, Education, User Level, and User Plus variables, which were defined in Section \ref{sec:data_description}. We also use the following  variables:
\begin{enumerate}
\item \textit{City Level:} JD has  classified customer cities   based on their most commonly used shipping address. City level takes on a value from   $\{$-1, 1, 2, 3, 4, 5$\}$, where level 1 corresponds to highly industrialized cities such as Beijing and Shanghai, level 2 cities correspond to provincial capitals,  and  level 3 to 5 cities are smaller cities. If there are no data,  then city level is set to -1. 
\item \textit{Purchase Power}: customer’s estimated purchase power indexed $\{$-1, 1, 2, 3, 4, 5$\}$, where -1 means no estimation, 1 represents the  highest,  and 5 represents  the lowest purchase power. 
\end{enumerate} 

\subsubsection{Product Variables:} We use \textit{Attribute 1} and \textit{Attribute 2}, which are  numerical integer variables between 1 and 4, and between 30 and 100, respectively. Both attributes characterize the functionality of a product, and a higher attribute value indicates better product performance.

\subsubsection{Seller Variables:} The following variables capture information about the seller. 
\begin{enumerate}
\item \textit{Product Type}: If the product is sold by JD, product type is  1. If the product is sold by a third-party  seller on JD, product type is 2. 
\item \textit{Original Price}: The original list price of the SKU before any discounts are applied. 
\item \textit{Promise}: shows the expected delivery time of the order in days. In the data, Promise could take a value from  $\{$N/A,1, 2, 3, 4, 5, 6, 7, 8$\}$, where  N/A is Not Available. 
\end{enumerate} 

\subsubsection{Promotion Variables:} These variables capture sellers' promotion strategies.
\begin{enumerate}
\item \textit{Number of Gifts}: the total number of free gift items that a customer receives in an order. This variable did not exist in the data set and we defined it  using the orders table. For each order ID, we identified the items that belonged to the order. If all items were free gifts, the order was eliminated from the analysis. Otherwise, we calculated the total number of free gifts in  the order and assigned this number to \textit{Number of Gifts} associated with the   item that the customer paid for (all orders that included some free gifts only had one paid item). We then erased the free gifts from the order. For example, if one order included SKU \#1 (paid), 2 units of SKU \#2 (free gift), and 2 units of SKU \#3 (free gifts), we set \textit{Number of Gifts} to 4 for SKU \#1, and deleted SKUs \#2 and \#3. The purpose of this variable is to capture the added value that a paid item generates for a customer when it comes with free gifts.  

\item \textit{Percentages of Direct Discount, Quantity Discount, Bundle Discount, and Coupon Discount per unit:} These variables show the value of each discount as a percentage  of the  original price.
In Python, the discount variables were coded  as ratios. One can  simply multiply the ratios by 100 to get percentages. 
\end{enumerate} 

\begin{remark}
\normalfont Initially, we wanted to use the total value (price) of the gift items as a predictive variable. However, the original   and final prices of gift items were both  recorded as zero in the data set. As a result, it was not possible to perform this analysis. 
\end{remark}
\begin{remark} 
\normalfont Before providing the data, JD had applied an allocation rule to calculate the contribution of each unit in an order to quantity   and bundle discounts, so we did not need to adjust the discount values for the order quantities. 
\end{remark}

\subsubsection{Other Variables:} These variables convey other aspects of orders. We use the variable \textit{Channel} from Section \ref{sec:data_description}. We also use categorical variables \textit{Day of Week} and \textit{Hour of Day}. 

 We used  LightGBM Regression to predict sales.  
Tabel \ref{tab:RMSE1}  reports the accuracy of LightGBM against alternative methods LASSO, Ridge, and KNN regression. 
LASSO regression uses the $\ell_1$ norm of the  regression coefficients to minimize overfitting. Ridge regression uses the squared $\ell_2$ norm for regularization. KNN regression uses the weighted average of the $k$ nearest points in the training set  to predict the response variable for a given observation. 
Accuracy  is measured by RMSE (Root Mean Squared Error). We see   that LightGBM regression performs significantly better. 

\begin{table}[tbh!]
    \centering
    \begin{tabular}{|l|c|c|c|c|}\hline
   \textbf{Method} &LightGBM Regression &LASSO Regression &Ridge Regression &KNN Regression\\\hline 
       \textbf{RMSE}  & 0.9367 &1.1512  & 1.1337  & 1.0130 \\\hline 
    \end{tabular}
    \vspace{0.3 cm}
    \caption{Benchmarking LightGBM Regression  against other   regression methods}
    \label{tab:RMSE1}
\end{table}

We now apply the SHAP method to interpret the results of the regression model and provide managerial insights to JD. 
To understand how SHAP values are calculated for a model, it is worth reviewing the concept of Shapley values. Consider a   cooperative game with $N$ players and pay-off function $v$ that maps subsets of players to real numbers. The Shapley value for player $i$ determines the contribution that the player brings to the game and is calculated as follows 
\begin{align}
\label{eq:shapleyvalue}
     \varphi_i(v) =  \sum_{S \subseteq N\setminus \{i\} }
           \frac{|S|!(N-|S|-1)!}{N!}
            \bigl( v(S\cup \{i\} ) - v(S) \bigr)
\end{align}
where $S$ is a subset of players of size $|S|$ that excludes player $i$.  Essentially, the formula calculates  the average additional value that player $i$ brings to coalition $S$ over all possible permutations.

Think of our ML model as a game, the predictions as the pay-off function, and the   variables as players.  Then Shapley values can be used to quantify the average contribution of each variable to the model. Note that the marginal contribution of variables to the model's prediction for an observation depends on the order in which the variables enter the model. In other words, the contribution of variable $i$ to the model depends on the set of variables $S$ that already exist in the model. Thus, the overall contribution of each variable to the model (global interpretability) is measured by the average   contribution (Shapley value) of the variable to each prediction.

\subsection{Global Interpretation} \label{subsec:analysis1-global}
\begin{figure}[h!]
    \centering
    \includegraphics[scale=0.6]{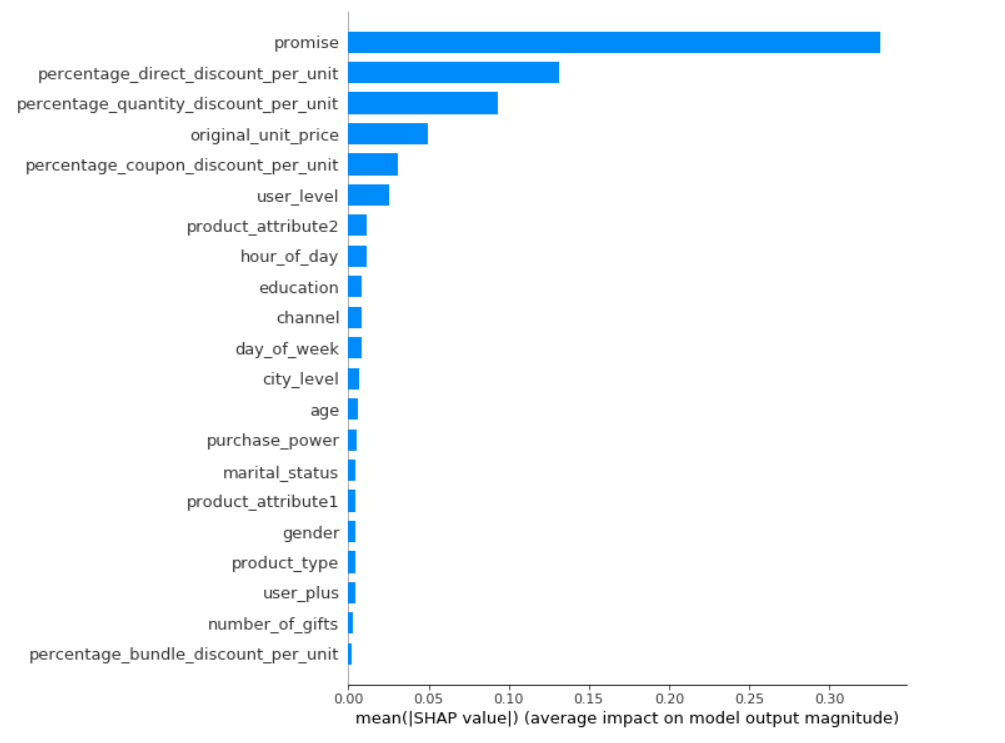}
    \caption{Variable importance plot for sales prediction }
    \label{fig:importance1}
\end{figure}

As the first step of SHAP analysis, we present the \textit{SHAP Variable Importance Plot}  in  Figure \ref{fig:importance1},  which illustrates
 the impact  of each variable on   the model's predictions.\footnotemark  \footnotetext{We would like to clarify that the word "impact" does not imply causality.}       Variable importance plots are useful in identifying and prioritizing the features that play a more important role in determining model outputs.  
The top five impactful variables for sales are: (1) expected delivery time in days (Promise); (2) per-unit  direct discount  as a percentage of the original  price; (3)  per-unit quantity discount  as a percentage of the original  price; 
(4) original price; 
(5) per-unit coupon discount  as a percentage of the original price. 

Naturally, the importance plots raise a follow-up question: How are the most impactful variables associated with the model's target variable?    \textit{SHAP Value Plots} 
can answer these questions.  The value   plots are made of  the training data and demonstrate the following information: 
(I) \textit{Impact:} The horizontal location shows whether the effect of that value is associated with a higher or lower prediction; (II) \textit{Original value:} Color shows whether that variable is high (in red) or low (in blue) for that observation; (III) \textit{Association:} Color and horizontal location of data points show whether a variable is positively or negatively correlated with the target variable. 

\begin{figure}[tbh!]
    \centering
    \includegraphics[scale=0.6]{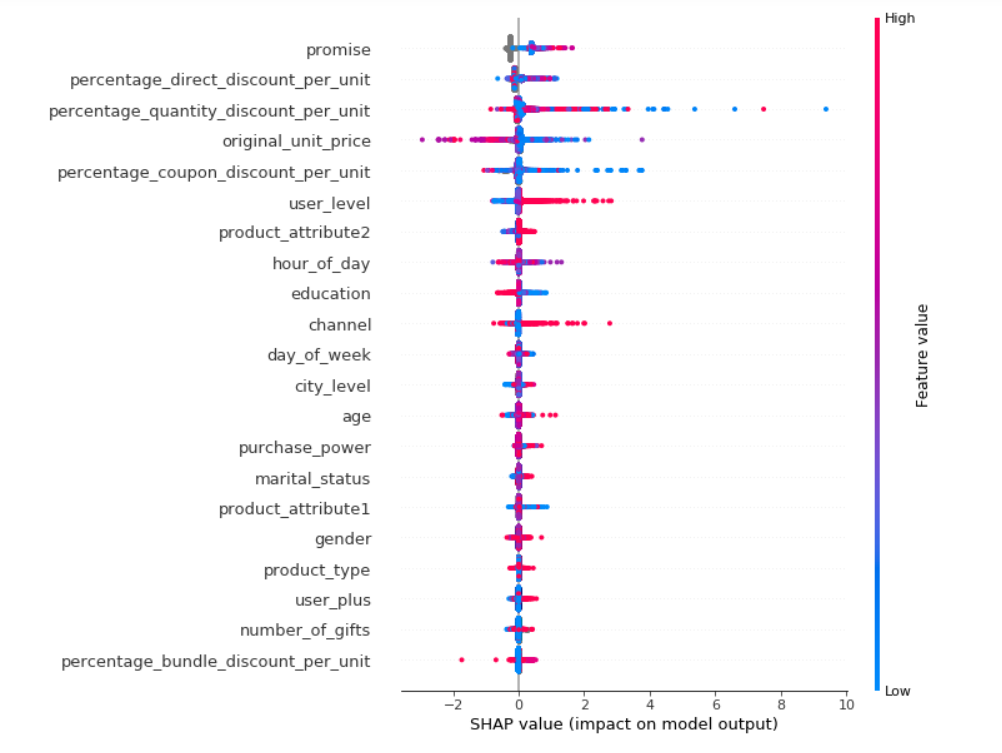}
    \caption{SHAP value plot for sales prediction }
    \label{fig:App-value1}
\end{figure}

\begin{figure}[h!]
     \centering
     \includegraphics[scale=0.75]{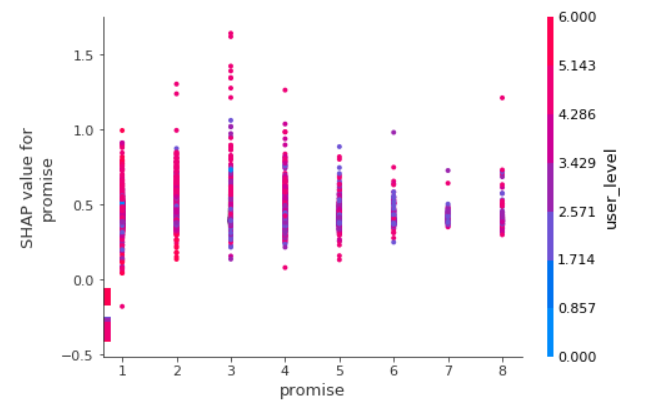}
    \caption{SHAP dependency plot for Promise. }
    \label{fig:promis-int}
\end{figure}

To understand how these top variables impact sales in detail, we can examine the SHAP dependency plots. First, we look at the   dependency plot  for Promise in Figure \ref{fig:promis-int}. It  illustrates how the SHAP value for Promise varies with Promise, with each dot representing an observation in the data. The vertical bar on the right hand side displays the variable in the model that Promise interacts with  most frequently using Friedman's H-statistic \cite{molnar2020interpretable, friedman2008predictive}. 

Let us  ignore the interaction aspect   for a moment (as if all the dots had the same color).  The plot shows that when Promise increases from one day, overall, there is an upward trend in customers' purchase volume. 
One plausible theory for this observation is customers' hoarding behavior. When delivery takes longer, customers tend to get more anxious as they need to spend more time and effort to monitor  their home inventories more frequently and  plan for the  subsequent orders  to avoid stockouts. To avoid this inconvenience, customers decide to increase their order size and stockpile larger inventories. The interaction visualization in Figure \ref{fig:promis-int} further supports our theory. User level has the strongest interaction with Promise, and   high  SHAP values for Promise belong to high user levels. In other words, as the promised delivery takes longer,  customers that historically make large purchases are the first to batch their orders. 

Such a batch-ordering behavior has been repeatedly documented in the supply chain literature through the famous Beer Game \cite{simchi2008designing}. When  lead times increase, supply chain  members use  order batching  to maintain their stocks. We conjecture that JD customers exhibit  similar behavior. Although customers' stockpiling behavior increases sellers' revenue in the short-term, we know from the literature that it leads to the bullwhip effect that bears many negative consequences such as poor customer service, poor capacity utilization, and unstable production schedules. Thus, our first recommendation to the sellers on JD.com is as follows.

\begin{Recommendation}
Given that the promised delivery is the most impactful feature for purchase volumes, JD sellers should achieve and maintain quick delivery to alleviate customers' batch ordering   and prevent the bullwhip effect. 
\end{Recommendation}

\begin{figure}[h!]
\begin{tikzpicture}
 \node[inner sep=0pt] at (0,0)
        {\includegraphics[width=.535\textwidth]{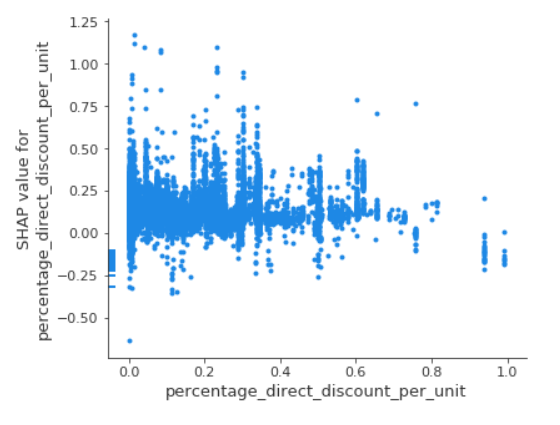}};
   \node[inner sep=0pt] at (0.14,-3.3)  {a)  };
\node[inner sep=0pt] at (8.3,0)
   {\includegraphics[width=.535\textwidth]{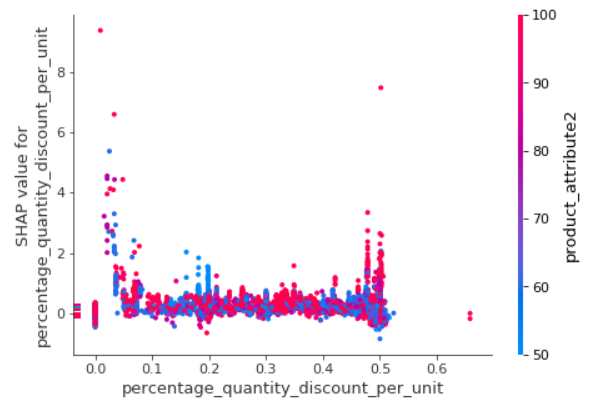}};
    \node[inner sep=0pt] at (8.3,-3.2) {b)};
\end{tikzpicture}
\begin{tikzpicture}
 \node[inner sep=0pt] at (0,-1)
        {\includegraphics[width=.535\textwidth]{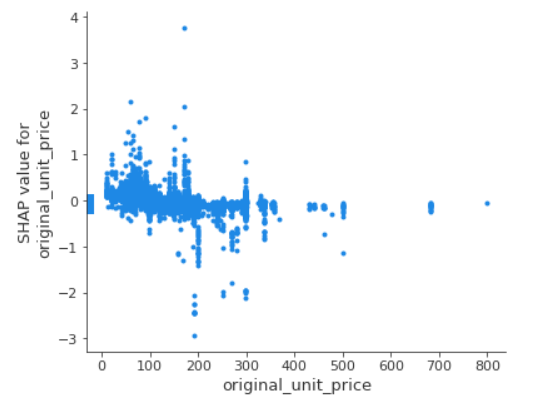}};
    \node[inner sep=0pt] at (0.14,-4.6)  {c)  };
\node[inner sep=0pt] at (8.3,-1.15)
   {\includegraphics[width=.55\textwidth]{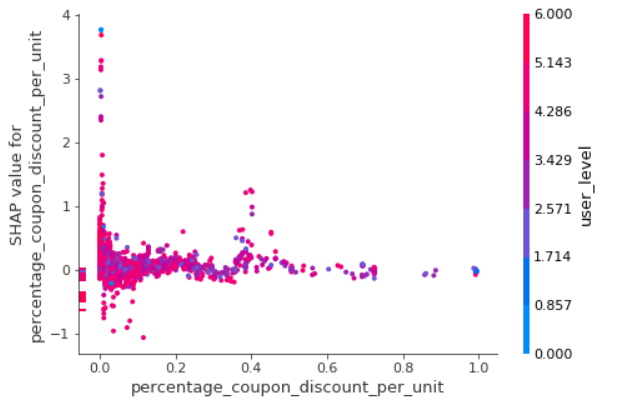}};
    \node[inner sep=0pt] at (8.3,-4.6) {d) };
\end{tikzpicture}
    \caption{SHAP dependency plots for direct, quantity, and coupon discounts, and for original price. } \label{fig:scatters-analysis1}
\end{figure}
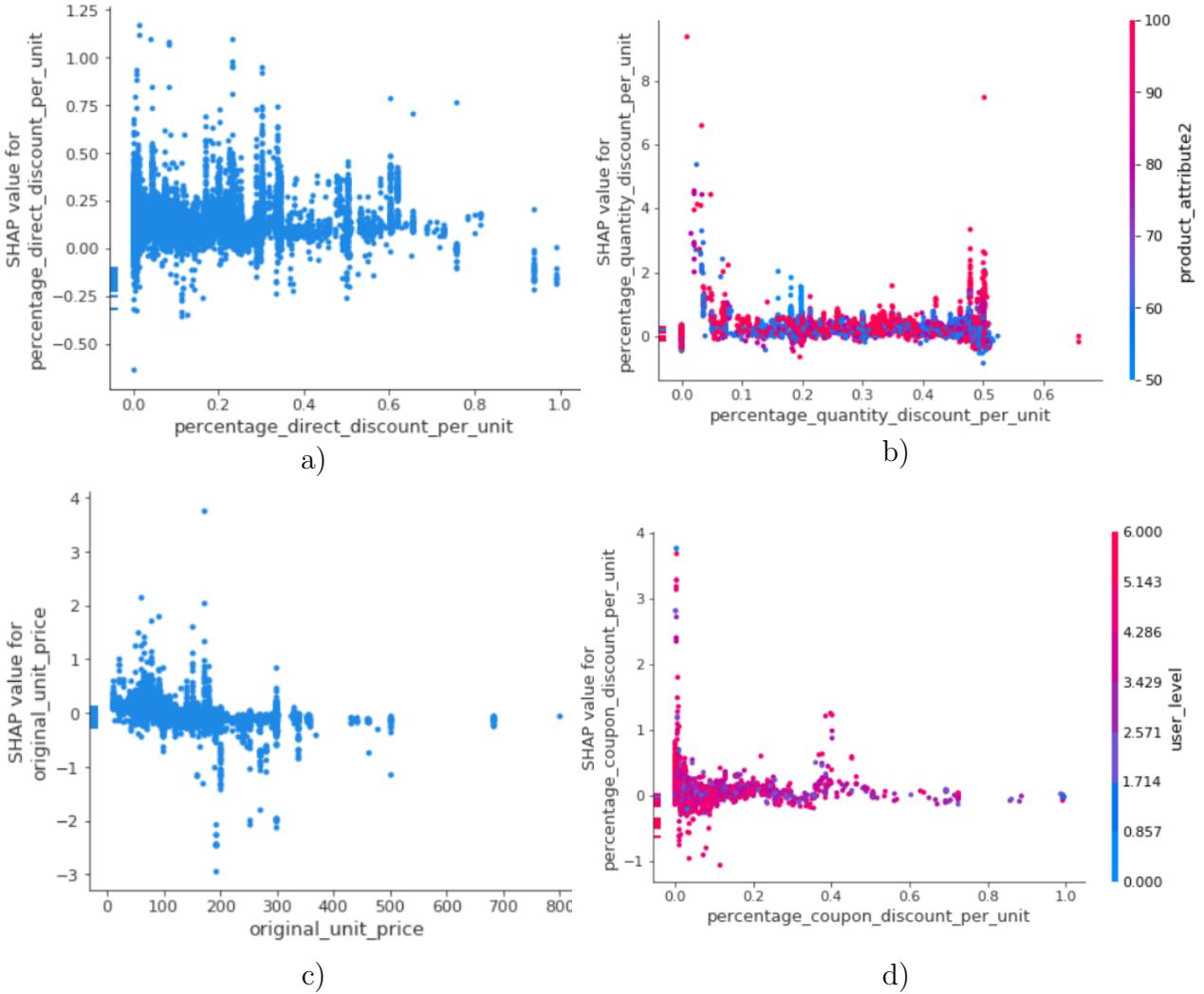

Next, we look at the SHAP dependency plot for  the per-unit direct discount as a percentage of the original price in Figure \ref{fig:scatters-analysis1}(a). We observe that direct discount has a positive impact on sales in general, even when the amount of discount is a small percentage of the listed price.

\begin{Recommendation}
Sellers should utilize the effectiveness of   direct discounts in boosting sales. A little direct discount goes a long way.
\end{Recommendation}

Now we examine Figure \ref{fig:scatters-analysis1}(b) for the effect of quantity discounts. The plot shows that quantity discounts are effective, even when they are a small fraction of the listed price. Also, we generally observe that higher values of Attribute 2 increase the positive impact of quantity discounts.   

\begin{Recommendation}
As the second most impactful promotion strategy, sellers should take advantage of quantity discounts. 
In particular, sellers should offer quantity discounts on SKUs whose Attribute 2 values are higher. 
\end{Recommendation}

Next, the value plot shows that, as expected, higher prices reduce sales. Figure \ref{fig:scatters-analysis1}(c) allows us to see this trend in detail.  The most salient observation in this plot is that the SHAP value for price turns negative near $200$ RMB. 

\begin{Recommendation}
List prices that exceed 200 RMB hinder sales, unless the seller offers discounts to reduce the net cost. In doing so, sellers should use the discount methods that are deemed effective in the SHAP analysis. 
\end{Recommendation}

Finally, Figure \ref{fig:scatters-analysis1}(d) shows the association between sales and coupon discount percentages. The dominant trend in the SHAP values and the interaction effect lead us to the next recommendation.

\begin{Recommendation}
In addition to direct and quantity discounts, sellers should use coupon discounts to increase sales. In particular, sellers should extend coupons to customers who have made larger total purchases in the past. 
\end{Recommendation}


\vspace{20pt}

\subsection{Local Interpretation} \label{subsec:analysis1-local}
In this section, we describe  SHAP's local interpretation  capabilities. This feature of SHAP helps model users understand how the prediction model arrives at a particular prediction and illustrates the contribution of each variable. 

\begin{figure}[h!]
  \centering
    \includegraphics[scale=0.6]{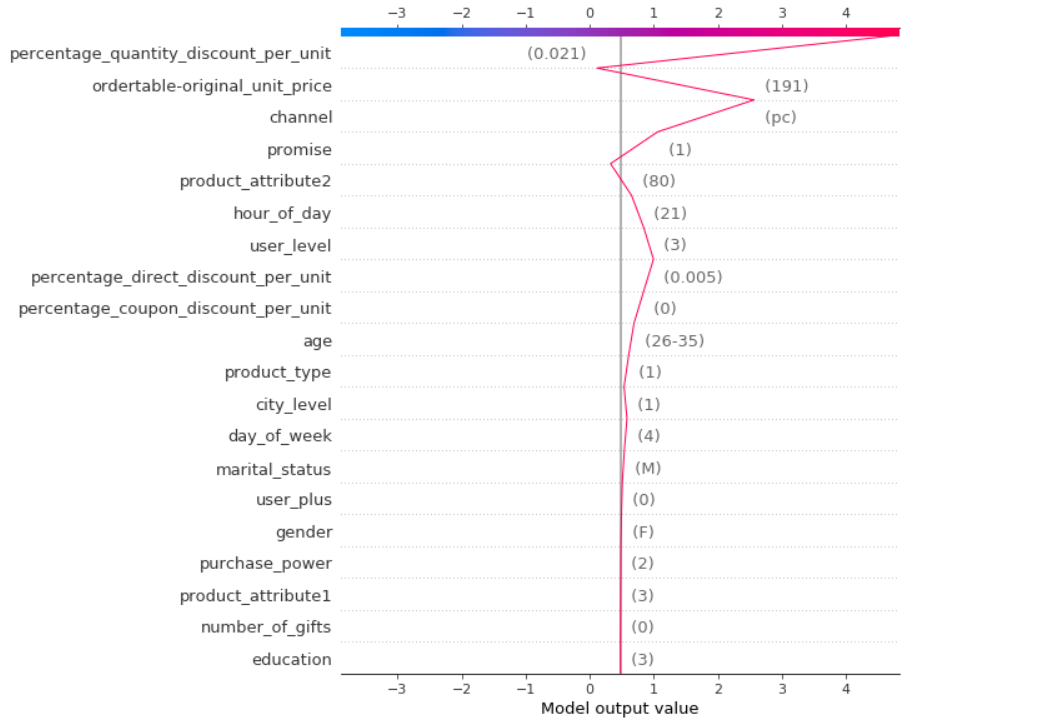}
    \caption{ Decision plot for a prediction. The base value is 0.4767. The sales prediction is 4.66}
    \label{fig:shapforeceplot1}
\end{figure}

\textit{SHAP Decision Plots} are  created to effectively explain a model's prediction for any observation.  Figure \ref{fig:shapforeceplot1} shows   an example of a  decision plot for our sales model. 
The  straight vertical line marks the model's base value. The colored line is the prediction. Feature values are printed next to the prediction line. 
The product has attributes are  3 and 80, is sold by JD at 191 RMB for next-day delivery, and does not come with any gifts. The unit discounts are 2.1\% (quantity discount) and 0.5\% (direct discount) off the list price. The customer is married and female, age 26-35, holds a Bachelor's degree, lives in a highly-industrialized  city, has a relatively high purchase power, is of user level 3, and is not a PLUS member. She is visiting JD through her PC at 9 p.m. on a Thursday.
Starting at the bottom, the prediction line shows how the SHAP values (i.e., the feature  effects) accumulate from the "base value" (0.4767) to reach the model's final prediction (4.66) at the top of the plot. The base value  marks the model's average prediction over the training set.


\section{Product Choice Analysis} \label{sec:class}
The purpose of this section is to build a classification model that would predict the probability that a customer would purchase a particular SKU. 
We use the same explanatory variables as  in Section \ref{sec:sales}, except  Attribute 1 and Attribute 2, which were used for clustering.

\subsection{Clustering SKUs}
We   filtered the 31,868 SKUs in the data set based on whether or not  their  Attribute 1 and  Attribute 2  values were recorded.
Only 13,725 SKUs had both attributes recorded.  We dropped the   18,143  SKUs whose Attribute 1 and/or Attribute 2 values were missing. 
We did not use imputation for two reasons. First, the number of SKUs with missing values  is much larger than the number of  SKUs with recorded values. Thus, any imputation method would entail significant errors and adversely impact  the reliability of our analysis. Second, the number of SKUs with available data  is   sufficiently large to conduct a meaningful study on customers' product choice behavior without requiring to estimate the missing data. 

We clustered the 13,725 SKUs into classes based on the two attributes, using  the $K$-means clustering algorithm \cite{lloyd1982least}. 
The  Elbow Method 
indicated that using four clusters was appropriate. Table \ref{tab:clusters}  shows the number of SKUs assigned to each cluster.  

\begin{table}[tbh!]
    \centering
    \begin{tabular}{|l|c|c|c|c| }\hline
  \textbf{Cluster} &0 &1  &2 & 3 \\\hline 
      \textbf{SKUs}  & 3754 & 3365  & 4685 & 1921
      \\\hline 
    \end{tabular}
    \vspace{0.3 cm}
    \caption{Number of SKUs assigned to each cluster}
    \label{tab:clusters}
\end{table}

The LightGBM  algorithm can be used for both classification and regression. Table \ref{tab:precision}  reports the precision of LightGBM against alternative classification methods.  The idea of Na\"ive Bayes is to calculate $P(\mathbf{X}|C_k)$ for each observation,  the  conditional probability  of a given set of feature values  $\mathbf{X}$, given the response value belongs to class $k$. Then each observation is assigned to the class with the highest $P(\mathbf{X}|C_k)$. Once the model is trained, for a new observation, these conditional probabilities are calculated for its $\mathbf{X}$ to classify the observation. However, calculating the conditional probabilities using   the Bayes' Rule requires  calculating joint probabilities that are difficult to estimate in practice.  Na\"ive Bayes  assumes that the elements of $\mathbf{X}$ are mutually independent to make the calculation of  $P(\mathbf{X}|C_k)$ easier. 
In the K-Nearest Neighbors (KNN) Classifier, an observation is classified by a majority vote of its neighbors, with the observation being assigned to the class most common amongst its $K$ nearest neighbors \cite{cover1967nearest, altman1992introduction}. 

\begin{table}[tbh!]
    \centering
    \begin{tabular}{|l|c|c|c| }\hline
  \textbf{Method} &LightGBM Classifier &Na\"ive Bayes Classifier &KNN Classifier\\\hline 
      \textbf{Precision}  & 0.8653 &0.3967  & 0.6920 \\\hline 
    \end{tabular}
    \vspace{0.3 cm}
    \caption{Performance of LightGBM classifier  against other   classification methods}
    \label{tab:precision}
\end{table}

\subsection{Global Interpretation}
\label{subsec:analysis2-global}
We now examine the   SHAP  importance,  value, and dependency plots in  Figures \ref{fig:importance2-first} to    \ref{fig:Appendix-dep3}.\footnotemark \footnotetext{Due to space limitations, we moved the plots to the Online Companion.} The importance plots show that, for all four classes, financial factors such as price and discount play a much more critical role in influencing customers’ choice behavior.

In interpreting the SHAP plots, one should differentiate between two different types of variables. The first group of variables includes those that characterize an incoming visit to the platform, but do not depend on the purchased product. We refer to these variables as customer-specific variables (such as education, channel, hour of the day, etc.). In contrast, other variables -- referred to as platform-specific variables -- vary by product and are set by the platform for each SKU before the   customer arrives (such as price, product type, promise, and various   discounts).

For the first variable group, any change in the variable can be directly linked to the likelihood of a purchase from one of the classes. For instance, one can infer whether a higher level of education (while keeping everything else constant) leads to a higher or lower chance of purchasing an SKU from class 2. However, the same conclusion does not hold about the second  variable group because these variables  correspond to many different SKUs that belong to a particular class. Therefore, any changes in the value of these variables and their corresponding SHAP values should be used to explain how customers are choosing different SKUs within that class. For instance, we may observe that among all SKUs that belong to a particular class, those with a higher price also have a higher probability of purchase (e.g., because customers prefer them for reasons other than price).

To draw insights about how customers’ choice behavior, we distinguish between these two variable groups and focus on each group's most influential variables separately for each of the four classes. Based on the SHAP plots, we present the following observations: 

\begin{observation}
Customer-specific variables seem to have low predictive power for all four classes. Therefore, the incoming customer's demographic information, the day and time of her visit, and the channel that she is using does not reveal a significant amount of information about the customers’ preference for one class over the other. Put differently, the probability of purchase from any of the four classes remain largely unaffected by customer-specific variables. The only exceptions to this statement is Education Level for Class 3 products and Day of the Week for Class 1 products. More precisely, keeping everything else constant, an incoming customer to JD.com is more likely to prefer an SKU from Class 3 over other classes if the customer has a lower education level. Similarly, a new visitor is more likely to prefer an SKU from Class 1 if her visit happens during earlier days of the week (keeping everything else constant). 
\end{observation}

This observation can be incorporated into JD’s demand forecast system and improve inventory management decisions, both temporally and spatially. Demographic information can also be used toward targeted marketing, and identifying customers that are more likely to be persuaded by personalized ads. The platform can utilize such information in its targeting ads and customize its product portfolio (i.e., the items shown to the customer in the first page upon their visit to the platform) according to   customers’ specific characteristics.

\begin{observation}
Among SKUs in Class 0 and Class 3, those with lower prices seem to be more popular than more expensive ones. This conclusion, however, does not hold about SKUs in Class 1 and Class 2. Customers’ preference in choosing between Class 1 and Class 2 products is mostly driven by factors other than price.
\end{observation}

\begin{observation}
Direct Discount seems to be effective for Class 0 items only if it is sufficiently high (i.e., higher than 40\%). However, the opposite is true for Class 3, so that items receiving a moderate level of Direct Discount are more popular with the customers who have purchased from this class. More interestingly, the low customer price sensitivity for Class 1 and Class 2 items is also reflected in their reaction to Direct Discount for these classes. Specifically, Direct Discount is largely ineffective in persuading customers to buy discounted items in Class 1 and Class 2. In fact, SKUs that have received large Direct Discount in Class 2 remain highly unpopular.
\end{observation}

\begin{observation}
Quantity Discount is an effective tool to influence purchase for Class 0 products and Class 3 products (if the discount is large enough). However, similar to Direct Discount, Quality Discount is not the main factor influencing customers’ preference for items that belong to Class 1 and Class 2.
\end{observation}

Based on this observation, JD should use Direct Discount and/or Quantity Discount to increase the demand for items in Class 0 and Class 3. These discounting methods, however, are mostly ineffective for Class 1 and Class 2 items  (For SKUs in Class 2, Bundle discount is the only effective discounting strategy, and hence should be preferred over Direct discounting and Quantity discounting). For   Class 1 and Class 2 items, customers’ preferences seem to be less driven by price and discount considerations. Therefore, targeted marketing is a more effective tool for these items compared to various forms of discount.

\begin{observation}
Customers who have purchased from Class 1, Class 2, and Class 3 are largely insensitive about Product Type. In other words, whether JD or a third party has fulfilled an order has not been a major driver for customers’ product choice. For Class 0, on the other hand, customers seem to have a strong preference for SKUs sold by third-party sellers. 
\end{observation}

\begin{observation}
Customers’ product choice is  largely unaffected by the promised delivery time. That is, customers are unlikely to switch their favorite product choice due to a longer delivery time.
\end{observation}

    \section{Discussion and Conclusion} \label{sec:conclusion}

We analyzed 454,897 JD buyer users who ordered at least one item from a particular product category in March 2018. Our descriptive statistical analysis 
reveals that $50\%$ of buyer users spent less than  $80$ RMB and around $90\%$ spent less than  $210$ RMB. 
Most buyer users (i.e., $99.46\%$) ordered either one or two units. While $40\%$, $54\%$ and $78\%$ of buyer users had less than $1$ hour, $1$ day, and $1$ week interaction with the site,  these percentages for non-buyer users grow significantly to $70\%$, $77\%$ and $88\%$. Thus, on average, buyer users spend more time than non-buyer users.

We also performed empirical statistical analyses of users' spending behavior 
More precisely, we constructed the empirical probability distribution of total spending, total discount, and number of units per order for different levels of six features consisting of age, gender, education, marital status, PLUS status, and user level. Our analysis illustrates that while the distribution of total spending and total discount for buyer users with respect to the features age, gender, education, and marital status appear similar, they could be different for the PLUS status and user level features. More precisely, while user behavior in different levels of the first four features did not vary dramatically, there were considerable changes in behavior at different levels of PLUS status and user level. However, for all given six features, most buyer users ordered only one unit at a time, which is consistent with the descriptive statistical output.

We also built  models to predict sales and customers' product choice. We applied a novel framework to interpret the predictions. 
Our findings lead to an important insight into the effect of delivery timing on customers’ purchase decisions. Specifically, even though customers’ product choice is highly insensitive to the promised delivery time, this factor significantly impacts customers’ order quantity. In other words, longer delivery times do not induce customers to switch their preferred product but persuade them to order more, which can potentially entail an undesirable bullwhip effect for the platform. Furthermore, the effectiveness of various discounting methods depends on the specific product and the magnitude of the discount. Therefore, JD should exercise carefully utilize the right method of discounting as well as its amount. For instance, we observe that direct discount and quantity discounts are highly ineffective for persuading customers to purchase from certain classes of products, while their effectiveness for other classes depends on the magnitude of the discount. When it comes to the purchase quantity (i.e., order size), even a small level of discount can have a substantial effect. Finally, 200 RMB seems to be an anchor point for customers who are buying this product category, and the shift in their behavior in the vicinity of this price point is the largest.

\section*{Declarations}

\begin{itemize}
\item Funding: Not applicable
\item Competing interests: Not applicable
\item Ethics approval: Not applicable
\item Consent to participate: Not applicable
\item Consent for publication: Not applicable
\item Availability of data and materials: Not applicable
\item Code availability: Not applicable
\item Authors' contributions: Not applicable
\end{itemize}

	
	


	\bibliographystyle{plain}
	\bibliography{Biblio}

\begin{thebibliography}{10}

\bibitem{Abolghasemi2020}
M.~Abolghasemi, J.~Hurley, A.~Eshragh, and B.~Fahimnia.
\newblock Demand forecasting in the presence of systematic events: Cases in
  capturing sales promotions.
\newblock {\em International Journal of Production Economics}, 230:107892,
  2020.

\bibitem{Alexandrov2019}
A.~Alexandrov, K.~Benidis, M.~Bohlke-Schneider, V.~Flunkert, J.~Gasthaus,
  T.~Januschowski, D.C. Maddix, S.S. Rangapuram, D.~Salinas, J.~Schulz,
  L.~Stella, A.C. T\"{u}rkmen, and Y.~Wang.
\newblock Probabilistic time series models in python.
\newblock {\em arXiv preprint arXiv:1906.05264}, 2019.

\bibitem{altman1992introduction}
N.S. Altman.
\newblock An introduction to kernel and nearest-neighbor nonparametric
  regression.
\newblock {\em The American Statistician}, 46(3):175--185, 1992.

\bibitem{Alvarado-Valencia2017}
J.~Alvarado-Valencia, L.H. Barrero, D.~\"{O}nkal, and J.T. Dennerlein.
\newblock Expertise, credibility of system forecasts and integration methods in
  judgmental demand forecasting.
\newblock {\em International Journal of Forecasting}, 33(1):298--313, 2017.

\bibitem{aouad2019market}
A.~Aouad, A.N. Elmachtoub, K.J. Ferreira, and R.~McNellis.
\newblock Market segmentation trees.
\newblock {\em arXiv preprint arXiv:1906.01174}, 2019.

\bibitem{Baecke2017}
P.~Baecke, S.~De~Baets, and K.~Vanderheyden.
\newblock Investigating the added value of integrating human judgement into
  statistical demand forecasting systems.
\newblock {\em International Journal of Production Economics}, 191:85--96,
  2017.

\bibitem{Bandara2019}
K.~Bandara, P.~Shi, C.~Bergmeir, H.~Hewamalage, Q.~Tran, and B.~Seaman.
\newblock Sales demand forecast in e-commerce using a long short-term memory
  neural network methodology.
\newblock In {\em Neural information processing}, pages 462--474, 2019.

\bibitem{Benidis2020}
K.~Benidis, S.S. Rangapuram, V.~Flunkert, B.~Wang, D.~Maddix, C.~Turkmen,
  J.~Gasthaus, M.~Bohlke-Schneider, D.~Salinas, L.~Stella, L.~Callot, and
  T.~Januschowski.
\newblock Neural forecasting: {I}ntroduction and literature overview.
\newblock {\em arXiv preprint arXiv:2004.10240}, 2020.

\bibitem{bravo2020mining}
F.~Bravo and Y.~Shaposhnik.
\newblock Mining optimal policies: A pattern recognition approach to model
  analysis.
\newblock {\em INFORMS Journal on Optimization}, 2(3):145--166, 2020.

\bibitem{Cha2004}
N.~Chakravarthy, A.~Spanias, L.D. Iasemidis, and K.~Tsakalis.
\newblock Autoregressive modeling and feature analysis of dna sequences.
\newblock {\em EURASIP Journal on Advances in Signal Processing}, 2004:13--28,
  2004.

\bibitem{Charkhgard2019}
H.~Charkhgard and A.~Eshragh.
\newblock A new approach to select the best subset of predictors in linear
  regression modelling: Bi-objective mixed integer linear programming.
\newblock {\em The ANZIAM Journal}, 61(1):64--75, 2019.

\bibitem{ciocan2020interpretable}
D.F. Ciocan and V.V. Mi{\v{s}}i{\'c}.
\newblock Interpretable optimal stopping.
\newblock {\em Management Science, Articles in Advance}, 2020.

\bibitem{cover1967nearest}
T.~Cover and P.~Hart.
\newblock Nearest neighbor pattern classification.
\newblock {\em IEEE transactions on information theory}, 13(1):21--27, 1967.

\bibitem{doshi2017towards}
F.~Doshi-Velez and B.~Kim.
\newblock Towards a rigorous science of interpretable machine learning.
\newblock {\em arXiv preprint arXiv:1702.08608}, 2017.

\bibitem{Eshragh2020}
A.~Eshragh, S.~Alizamir, P.~Howley, and E.~Stojanovski.
\newblock Modeling the dynamics of the {COVID-19} population in {A}ustralia:
  {A} probabilistic analysis.
\newblock {\em PLoS ONE}, 15(10):e0240153, 2020.

\bibitem{Eshragh2019}
A.~Eshragh, B.~Ganim, T.~Perkins, and K.~Bandara.
\newblock The importance of environmental factors in forecasting australian
  power demand.
\newblock {\em Environmental Modeling \& Assessment}, 27:1--11, 2022.

\bibitem{Eshragh2021}
A.~Eshragh, G.~Livingston, T.M.C. McCann, and L.~Yerbury.
\newblock {Rollage: E}fficient rolling average algorithm to estimate {ARMA}
  models for big time series data.
\newblock {\em arXiv preprint arXiv:2103.09175}, 2021.

\bibitem{Eshragh2022}
A.~Eshragh, F.~Roosta, A.~Nazari, and M.W. Mahoney.
\newblock {LSAR: E}fficient leverage score sampling algorithm for the analysis
  of big time series data.
\newblock {\em Journal of Machine Learning Research}, 23:1--36, 2022.

\bibitem{Fildes2009}
R.~Fildes, P.~Goodwin, M.~Lawrence, and K.~Nikolopoulos.
\newblock Effective forecasting and judgmental adjustments: an empirical
  evaluation and strategies for improvement in supply-chain planning.
\newblock {\em International Journal of Forecasting}, 25(1):3--23, 2009.

\bibitem{Fildes2019}
R.~Fildes, P.~Goodwin, and D.~\"{O}nkal.
\newblock Use and misuse of information in supply chain forecasting of
  promotion effects.
\newblock {\em International Journal of Forecasting}, 35(1):144--156, 2019.

\bibitem{friedman2008predictive}
J.H. Friedman, B.E. Popescu, et~al.
\newblock Predictive learning via rule ensembles.
\newblock {\em The Annals of Applied Statistics}, 2(3):916--954, 2008.

\bibitem{goodman2017european}
B.~Goodman and S.~Flaxman.
\newblock European union regulations on algorithmic decision-making and a
  “right to explanation”.
\newblock {\em AI magazine}, 38(3):50--57, 2017.

\bibitem{Goodwin2002}
P.~Goodwin.
\newblock Integrating management judgment and statistical methods to improve
  short-term forecasts.
\newblock {\em Omega}, 30(2):127--135, 2002.

\bibitem{Hamilton1989}
J.D. Hamilton.
\newblock A new approach to the economic analysis of nonstationary time series
  and the business cycle.
\newblock {\em Econometrica}, 57(2):357--384, 1989.

\bibitem{Hastie2009}
T.~Hastie, R.~Tibshirani, and J.~Friedman.
\newblock {\em The Elements of Statistical Learning: Data Mining, Inference,
  and Prediction}.
\newblock Springer, 2009.

\bibitem{Hewamalage2020}
H.~Hewamalage, C.~Bergmeir, and K.~Bandara.
\newblock Recurrent neural networks for time series forecasting: Current status
  and future directions.
\newblock {\em International Journal of Forecasting}, To Appear, 2020.

\bibitem{Hyndman2008}
R.J. Hyndman, A.B. Koehler, J.K. Ord, and R.D. Snyder.
\newblock {\em Forecasting with exponential smoothing: the state space
  approach}.
\newblock Springer Science \& Business Media, 2008.

\bibitem{Kremer2015}
M.~Kremer, E.~Siemsen, and D.J. Thomas.
\newblock The sum and its parts: Judgmental hierarchical forecasting.
\newblock {\em Management Science}, 62(9):2745--2764, 2015.

\bibitem{Lawrence2006}
M.~Lawrence, P.~Goodwin, M.~O'Connor, and D.~{\"O}nkal.
\newblock Judgmental forecasting: A review of progress over the last 25 years.
\newblock {\em International Journal of Forecasting}, 22(3):493--518, 2006.

\bibitem{Lawrence2000}
M.~Lawrence, M.~O'connor, and B.~Edmundson.
\newblock A field study of sales forecasting accuracy and processes.
\newblock {\em European Journal of Operational Research}, 122(1):151--160,
  2000.

\bibitem{lloyd1982least}
S.~Lloyd.
\newblock Least squares quantization in pcm.
\newblock {\em IEEE transactions on information theory}, 28(2):129--137, 1982.

\bibitem{Lundberg2017}
S.M. Lundberg and S.~Lee.
\newblock A unified approach to interpreting model predictions.
\newblock {\em Advances in neural information processing systems}, pages
  4765--4774, 2017.

\bibitem{Messner2019}
J.W. Messner and P.~Pinson.
\newblock Online adaptive lasso estimation in vector autoregressive models for
  high dimensional wind power forecasting.
\newblock {\em International Journal of Forecasting}, 35(4):1485--1498, 2019.

\bibitem{mivsic2020data}
V.V. Mi{\v{s}}i{\'c} and G.~Perakis.
\newblock Data analytics in operations management: A review.
\newblock {\em Manufacturing \& Service Operations Management}, 22(1):158--169,
  2020.

\bibitem{molnar2020interpretable}
C.~Molnar.
\newblock {\em Interpretable Machine Learning}.
\newblock Lulu. com, 2020.

\bibitem{Moritz2014}
B.~Moritz, E.~Siemsen, and M.~Kremer.
\newblock Judgmental forecasting: Cognitive reflection and decision speed.
\newblock {\em Production and Operations Management}, 23(7):1146--1160, 2014.

\bibitem{Mukherjee2018}
S.~Mukherjee, D.~Shankar, A.~Ghosh, N.~Tathawadekar, P.~Kompalli, S.~Sarawagi,
  and C.~Krishnendu.
\newblock {ARMDN: A}ssociative and recurrent mixture density net- works for
  retail demand forecasting.
\newblock {\em arXiv preprint arXiv:1803.03800}, 2018.

\bibitem{Salinas2020}
D.~Salinas, V.~Flunkert, J.~Gasthaus, and T.~Januschowski.
\newblock {D}eep{AR}: {P}robabilistic forecasting with autoregressive recurrent
  networks.
\newblock {\em International Journal of Forecasting}, 36(3):1181--1191, 2020.

\bibitem{Sanders2003}
N.R. Sanders and K.B. Manrodt.
\newblock The efficacy of using judgmental versus quantitative forecasting
  methods in practice.
\newblock {\em Omega}, 31(6):511--522, 2003.

\bibitem{shapley1953value}
L.S. Shapley.
\newblock A value for n-person games.
\newblock {\em Contributions to the Theory of Games}, 2(28):307--317, 1953.

\bibitem{She2018}
X.~Shen and Q.~Lu.
\newblock Joint analysis of genetic and epigenetic data using a conditional
  autoregressive model.
\newblock {\em BMC Genetics}, 16(Suppl 1):51--54, 2018.

\bibitem{simchi2008designing}
D.~Simchi-Levi, P.~Kaminsky, E.~Simchi-Levi, and R.~Shankar.
\newblock {\em Designing and managing the supply chain: concepts, strategies
  and case studies}.
\newblock Tata McGraw-Hill Education, 2008.

\bibitem{Trapero2015}
J.R. Trapero, N.~Kourentzes, and R.~Fildes.
\newblock On the identification of sales forecasting models in the presence of
  promotions.
\newblock {\em Journal of the operational Research Society}, 66(2):299--307,
  2015.

\bibitem{Trapero2013}
J.R. Trapero, D.J. Pedregal, R.~Fildes, and N.~Kourentzes.
\newblock Analysis of judgmental adjustments in the presence of promotions.
\newblock {\em International Journal of Forecasting}, 29(2):234--243, 2013.

\bibitem{Tversky1974}
A.~Tversky and D.~Kahneman.
\newblock Judgment under uncertainty: Heuristics and biases.
\newblock {\em Science}, 185:1124--1131, 1974.

\end{thebibliography}


	






\newpage
\onecolumn
\appendix
\allowdisplaybreaks

\section{Online Appendix: Supplementary Plots}
\label{Appendix:plots}


\begin{figure}[h!]
    \centering
    \includegraphics[angle=0, scale=0.6]{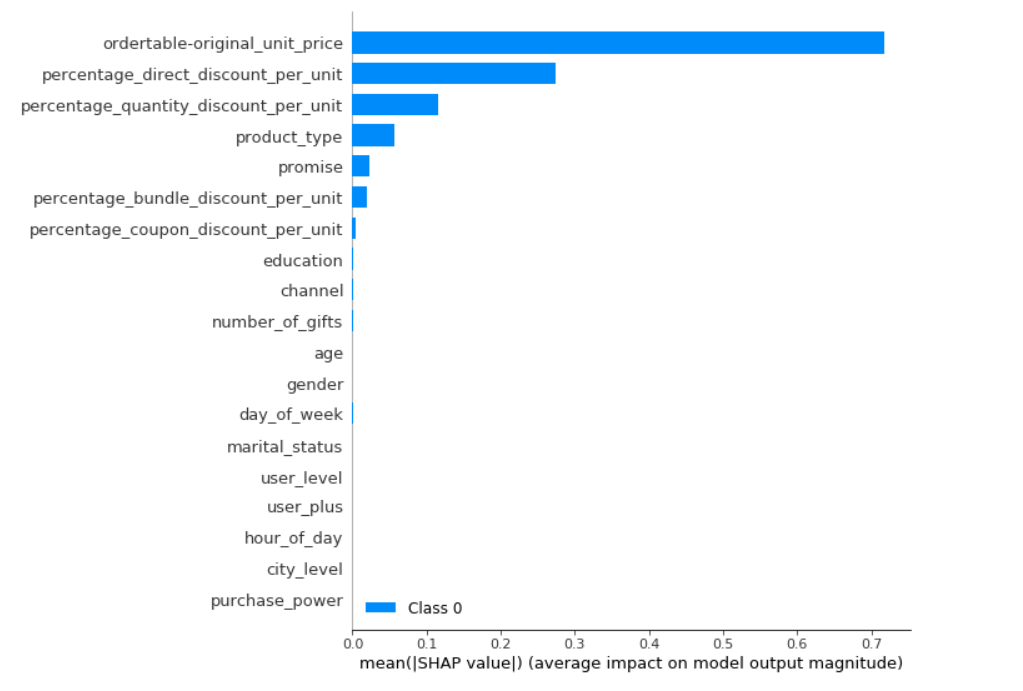}
    \includegraphics[angle=0, scale=0.6]{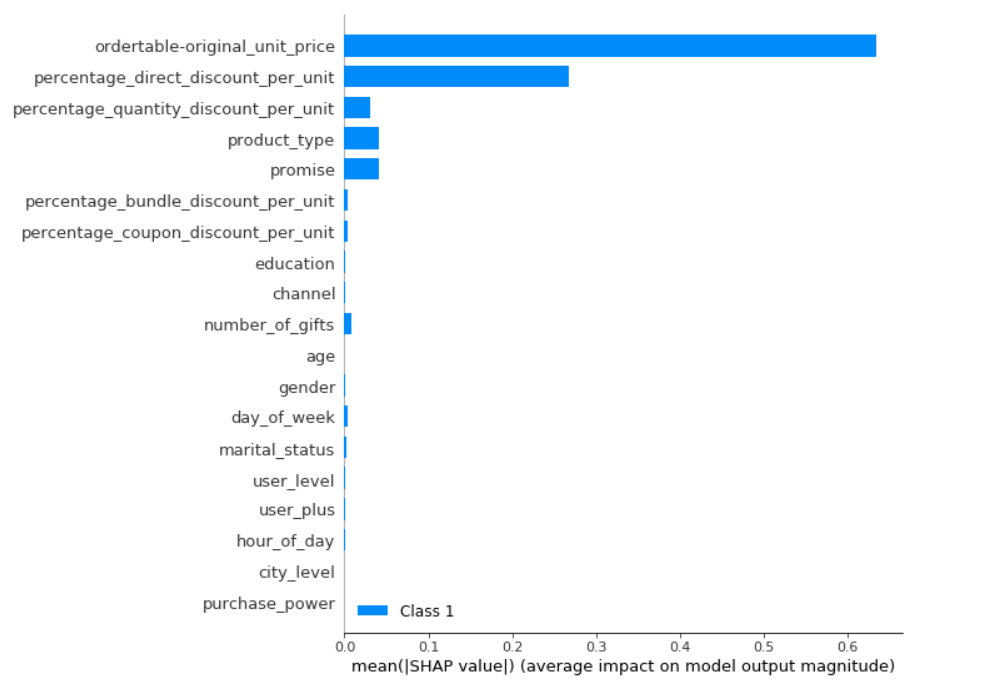}
        \caption{Variable importance plots for classes 0 and 1.}
        \label{fig:importance2-first}
\end{figure}
\begin{figure}[h!]
    \centering
    \includegraphics[angle=0, scale=0.65]{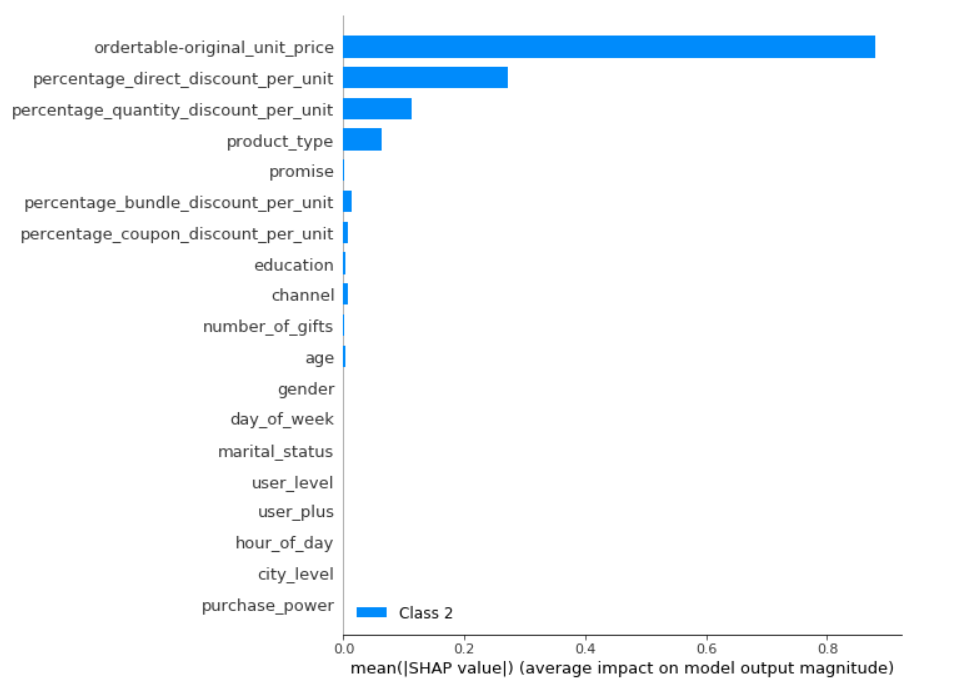}
    \includegraphics[angle=0, scale=0.65]{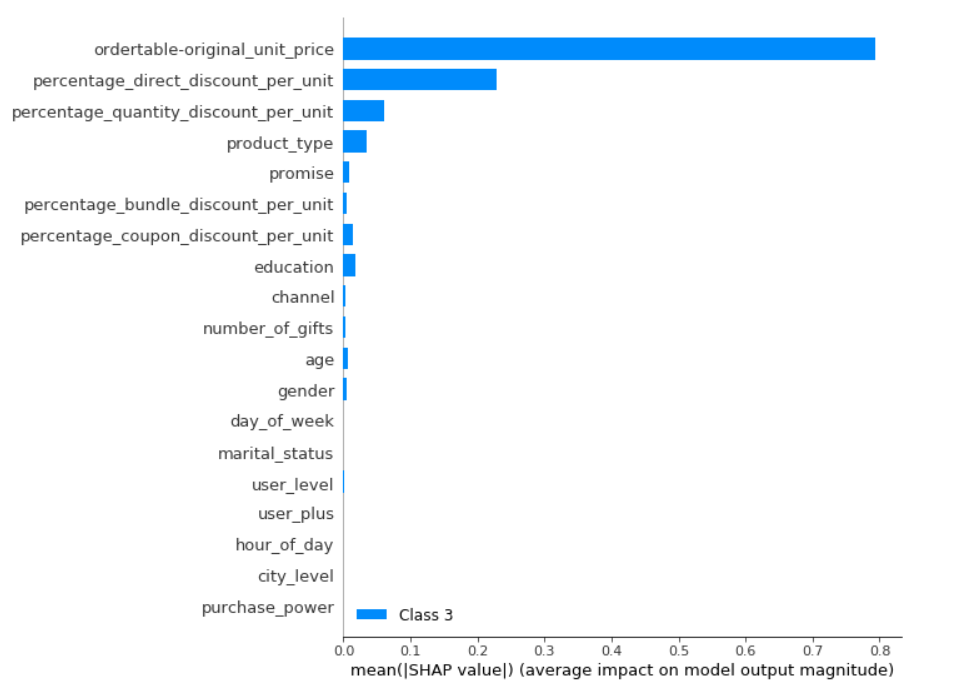}
    \caption{Variable importance plots for classes 2 and 3.}
    \label{fig:importance2-second}
\end{figure}

\begin{figure}[h!]
    \centering
    \includegraphics[angle=0, scale=0.62]{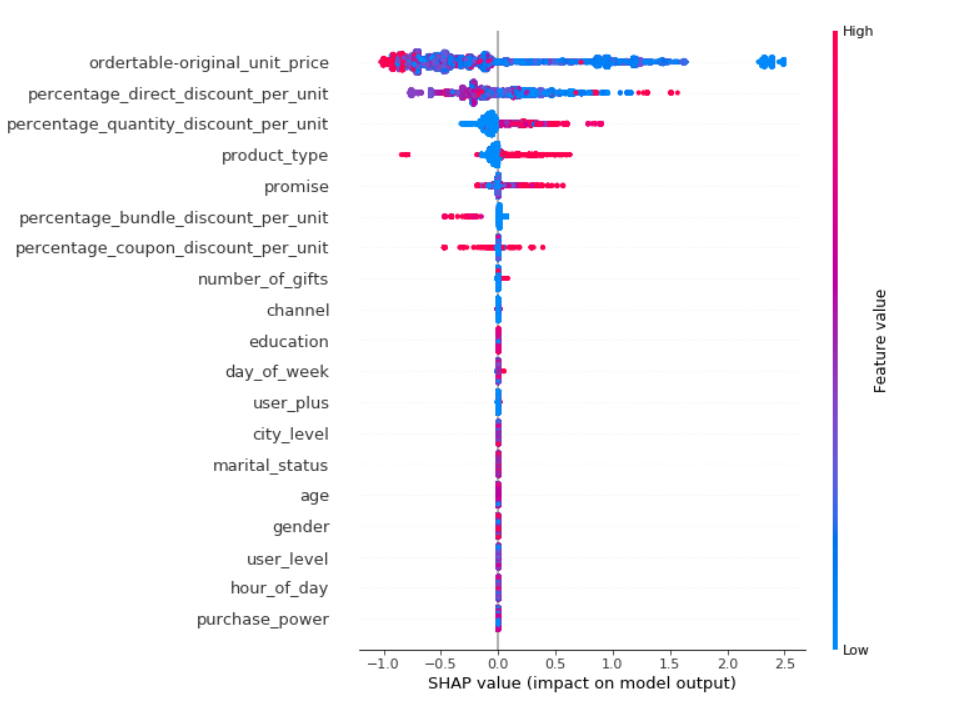}
    \includegraphics[angle=0, scale=0.62]{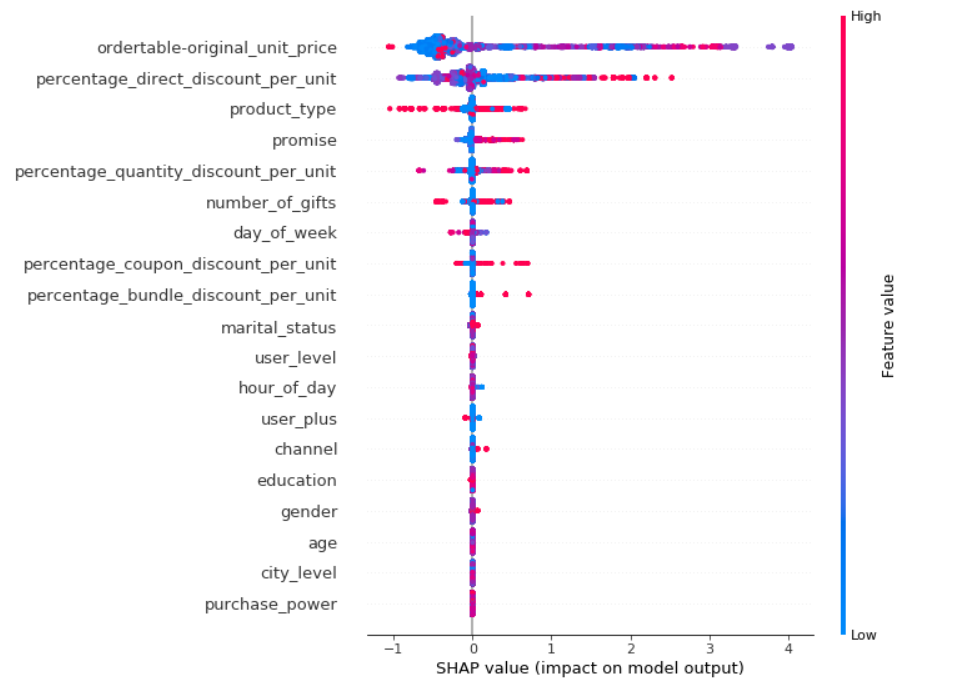}
        \caption{SHAP value plots for classes 0 and 1.}
        \label{fig:correlation2-first}
\end{figure}
\begin{figure}[h!]
    \centering
    \includegraphics[angle=0, scale=0.62]{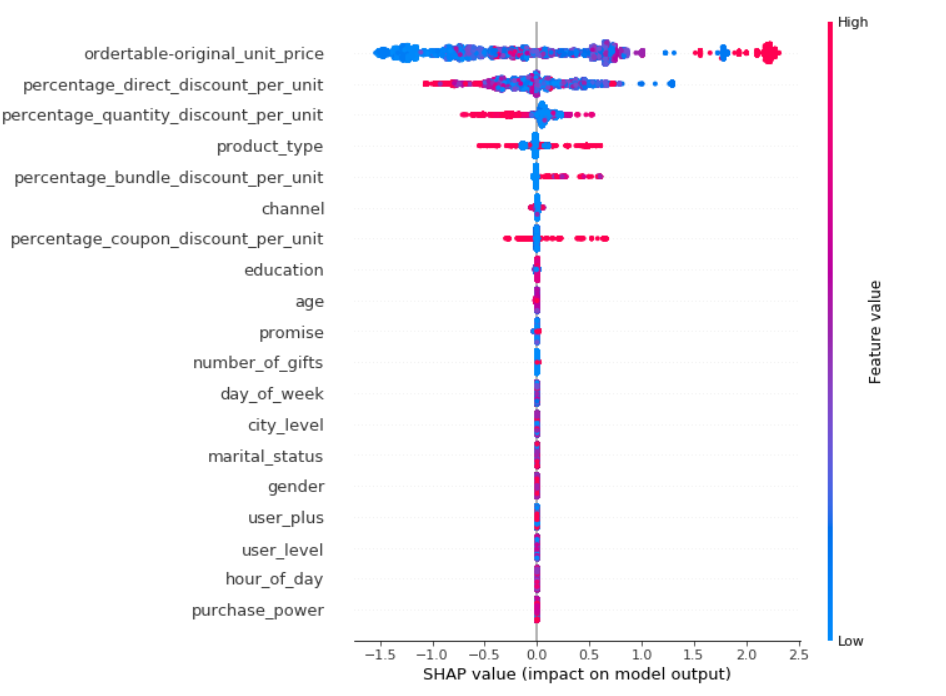}
    \includegraphics[angle=0, scale=0.62]{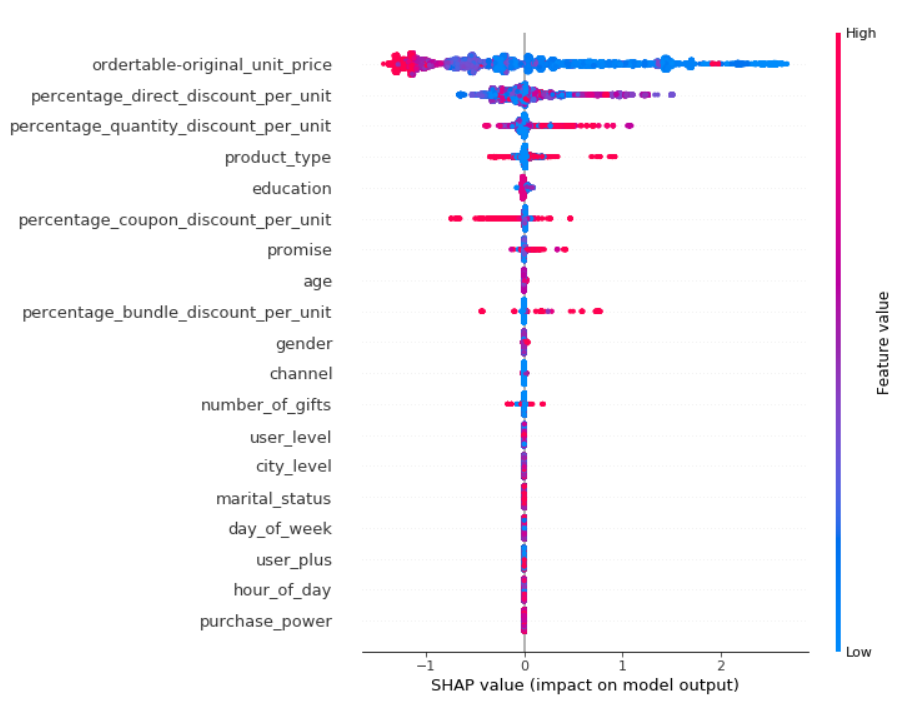}
    \caption{SHAP value plots for classes 2 and 3.}
    \label{fig:correlation2-second}
\end{figure}


\begin{figure}[h!]
\begin{tikzpicture}
 \node[inner sep=0pt] at (0,0)
        {\includegraphics[scale=0.45]{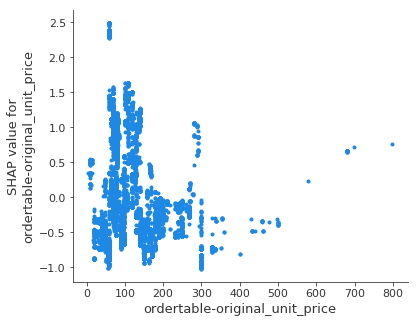}};
\node[inner sep=0pt] at (8.3,0)
  {\includegraphics[scale=0.45]{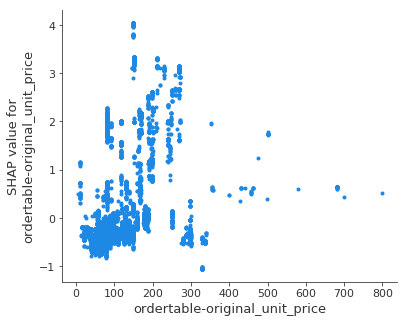}};
\end{tikzpicture}
\begin{tikzpicture}
 \node[inner sep=0pt] at (0,-1)
        {\includegraphics[scale=0.45]{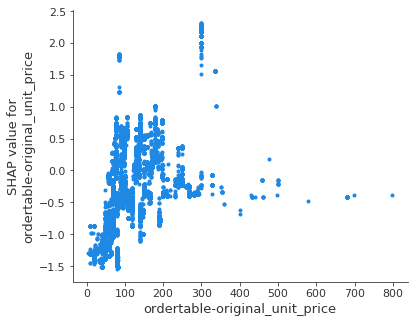}};
\node[inner sep=0pt] at (8.3,-1.15)
  {\includegraphics[scale=0.45]{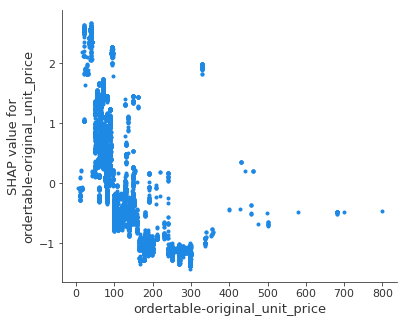}};
\end{tikzpicture}
\begin{tikzpicture}
 \node[inner sep=0pt] at (0,0)
        {\includegraphics[scale=0.45]{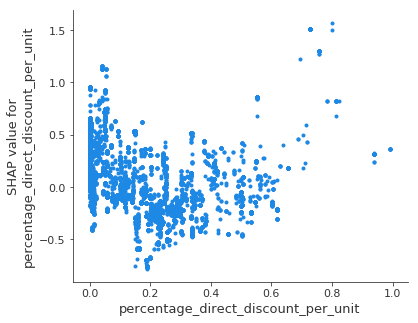}};
\node[inner sep=0pt] at (8.3,0)
  {\includegraphics[scale=0.45]{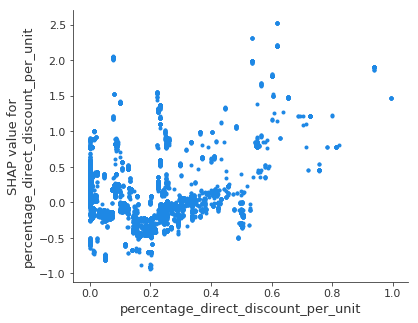}};
\end{tikzpicture}
\begin{tikzpicture}
 \node[inner sep=0pt] at (0,-1)
        {\includegraphics[scale=0.45]{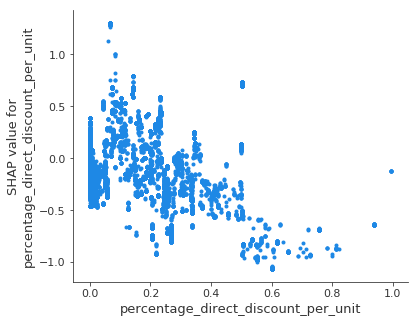}};
\node[inner sep=0pt] at (8.3,-1.15)
  {\includegraphics[scale=0.45]{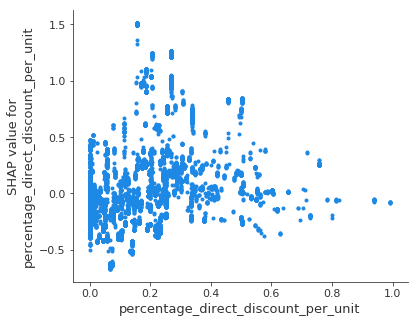}};
\end{tikzpicture}
    \caption{SHAP dependency plots for price (the first four) and direct discount (the second four).  Classes 0 to 3 are placed clockwise.} \label{fig:Appendix-dep1}
\end{figure}

\begin{figure}[h!]
\begin{tikzpicture}
 \node[inner sep=0pt] at (0,0)
        {\includegraphics[scale=0.45]{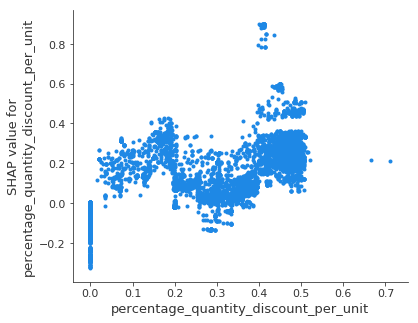}};
\node[inner sep=0pt] at (8.3,0)
  {\includegraphics[scale=0.45]{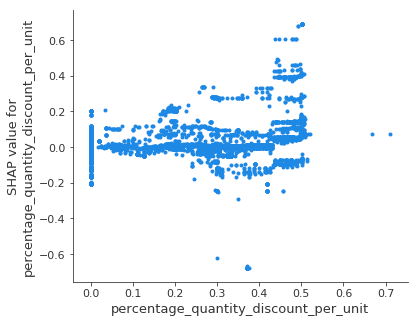}};
\end{tikzpicture}
\begin{tikzpicture}
 \node[inner sep=0pt] at (0,-1)
        {\includegraphics[scale=0.45]{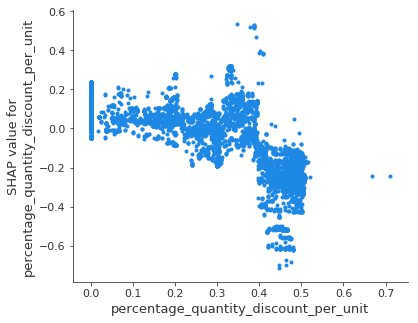}};
\node[inner sep=0pt] at (8.3,-1.15)
  {\includegraphics[scale=0.45]{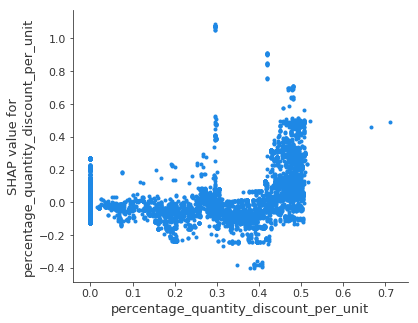}};
\end{tikzpicture}
\begin{tikzpicture}
 \node[inner sep=0pt] at (0,0)
        {\includegraphics[scale=0.45]{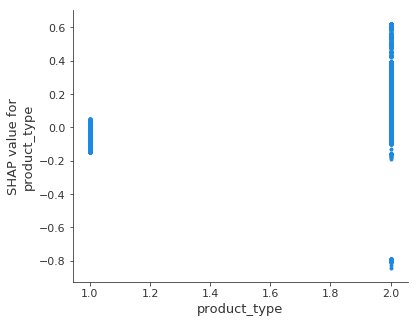}};
\node[inner sep=0pt] at (8.3,0)
  {\includegraphics[scale=0.45]{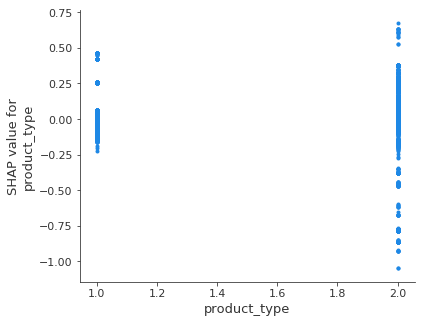}};
\end{tikzpicture}
\begin{tikzpicture}
 \node[inner sep=0pt] at (0,-1)
        {\includegraphics[scale=0.45]{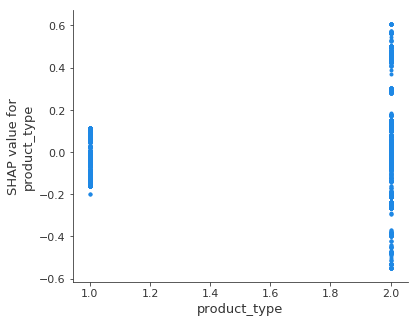}};
\node[inner sep=0pt] at (8.3,-1.15)
  {\includegraphics[scale=0.45]{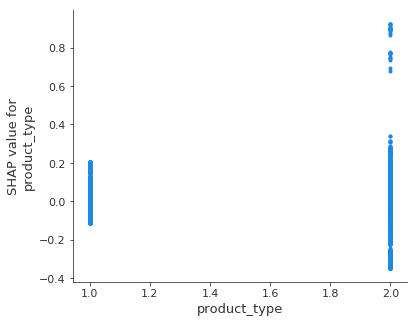}};
\end{tikzpicture}
    \caption{SHAP dependency plots for quantity discount (the first four) and product type (the second four).  Classes 0 to 3 are placed clockwise.} \label{fig:Appendix-dep2}
\end{figure}

\begin{figure}[h!]
\begin{tikzpicture}
 \node[inner sep=0pt] at (0,0)
        {\includegraphics[scale=0.45]{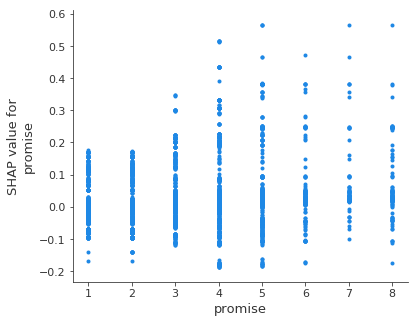}};
\node[inner sep=0pt] at (8.3,0)
  {\includegraphics[scale=0.45]{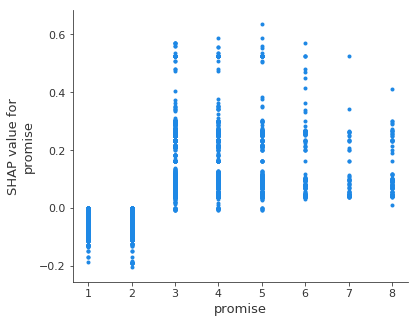}};
\end{tikzpicture}
\begin{tikzpicture}
 \node[inner sep=0pt] at (0,-1)
        {\includegraphics[scale=0.45]{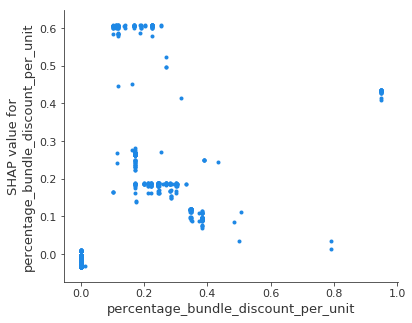}};
\node[inner sep=0pt] at (8.3,-1.15)
  {\includegraphics[scale=0.45]{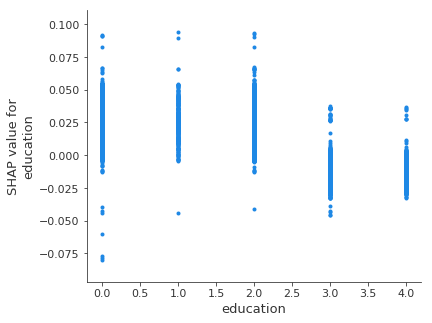}};
\end{tikzpicture}
        \caption{SHAP dependency plots for Promise (classes 0 and 1 in the first row), Bundle discount (class 2, bottom left), and Education (class 3, bottom right).} \label{fig:Appendix-dep3}
\end{figure}

\begin{figure}[h!]
    \centering
    \includegraphics[angle=0, scale=0.6]{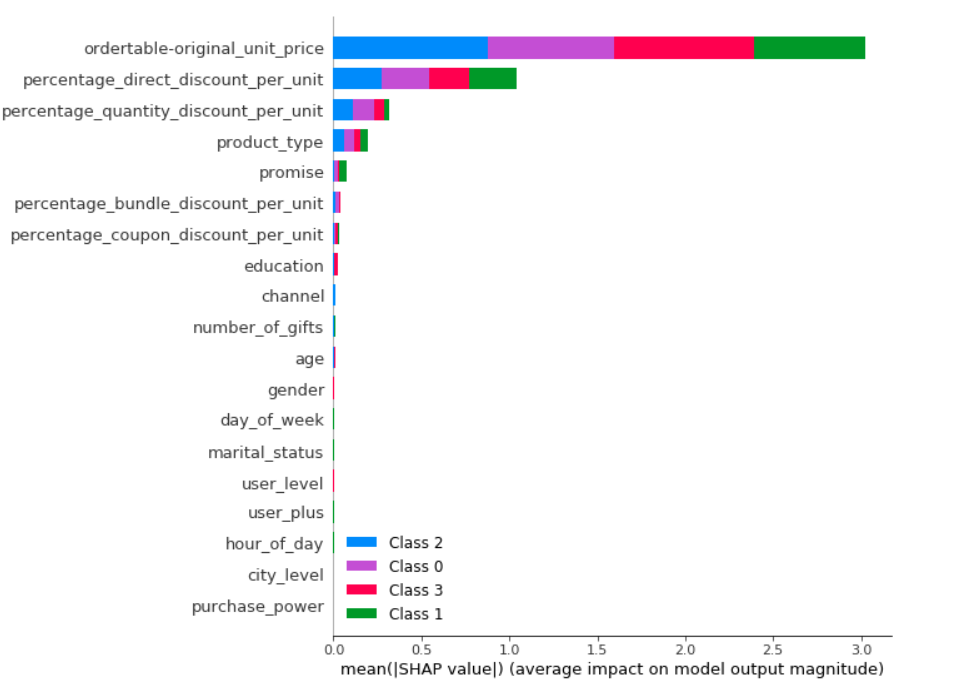}
    \caption{Aggregate variable importance plot for the classification model}
\label{fig:App-importance2-allclasses}
\end{figure}

\begin{figure}[tbh!]
    \centering
        \includegraphics[scale=0.65]{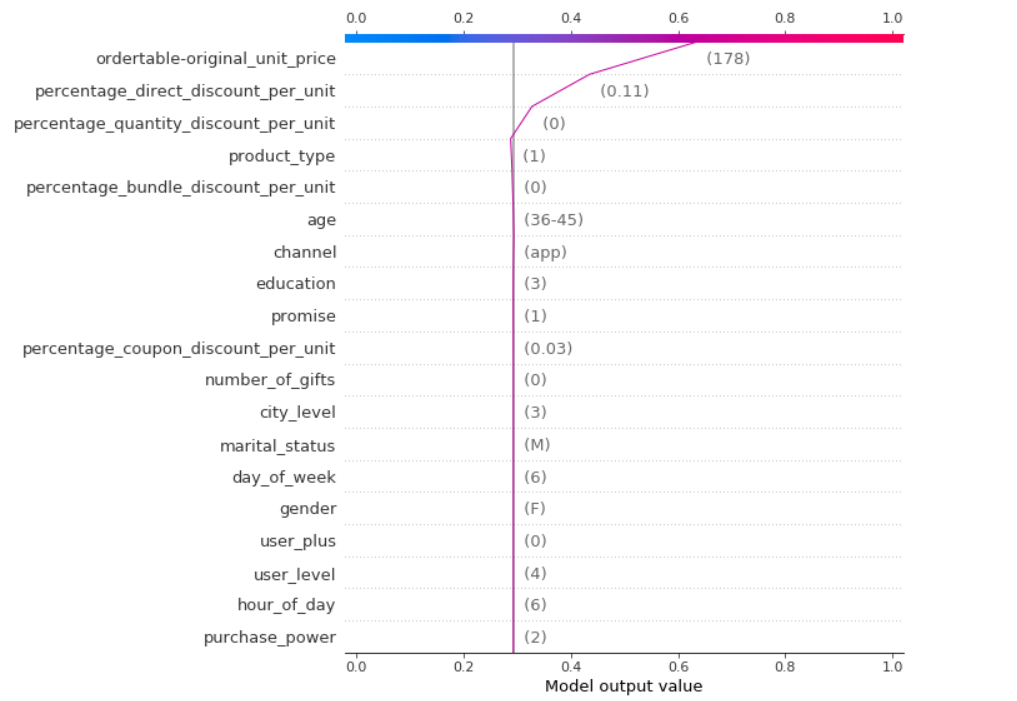}
    \caption{A Decision Plot  for     the classification analysis. The predicted probability is 0.63. }
     \label{fig:shapforeceplot2}
\end{figure}

\end{document}